\documentclass{article}
\usepackage{arxiv}

\usepackage[T1]{fontenc}
\usepackage{graphicx}
\usepackage{hyperref}

\usepackage{caption} 
\usepackage{subcaption}
\usepackage{todonotes}
\usepackage{multirow}
\usepackage{booktabs}
\usepackage{amsmath}
\usepackage{placeins}
\usepackage{float}
\usepackage{comment}

\usepackage{booktabs}
\usepackage{array}
\usepackage{ragged2e}
\newcolumntype{P}[1]{>{\RaggedRight\arraybackslash}p{#1}}

\usepackage{marginnote}

\NewDocumentCommand{\orcid}{m}{%
  \href{https://orcid.org/#1}{%
    \includegraphics[height=1.6ex]{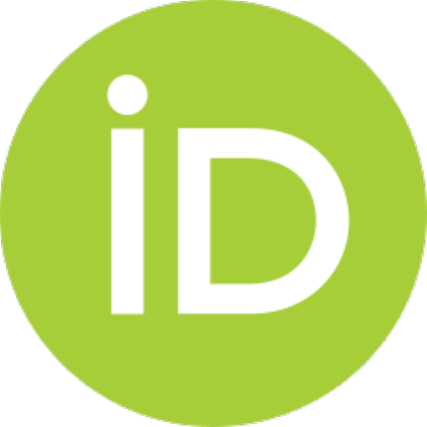}%
  }%
}

\NewDocumentCommand{\mail}{m}{%
  \href{mailto:#1}{\texttt{#1}}%
}

\title{SeNeDiF-OOD: Semantic Nested Dichotomy Fusion for Out-of-Distribution Detection Methodology in Open-World Classification. A Case Study on Monument Style Classification}

\author{\orcid{0009-0006-5843-3468}
    Ignacio Antequera-Sánchez$^{*,a}$ \\
    \mail{ignacioantequera@ugr.es}
    \And
    \orcid{0000-0001-8587-4345}
    Juan Luis Suárez-Díaz$^{a}$ \\
    \mail{jlsuarezdiaz@decsai.ugr.es}
    \And
    \orcid{0000-0002-0183-044X}
    Rosana Montes$^{b}$ \\
    \mail{rosana@ugr.es}
    \And
    \orcid{0000-0002-7283-312X}
    Francisco Herrera$^{a}$ \\
    \mail{herrera@decsai.ugr.es}%
}

\newcommand{\affiliations}{
  \vspace{-3em}
  \begin{center}
    \textsuperscript{*} Corresponding author \\
    \textsuperscript{a} \href{https://decsai.ugr.es}{Department of Computer Science and Artificial Intelligence (DECSAI)}, \href{https://dasci.es}{Andalusian Institute of Data Science and Computational Intelligence (DaSCI)}, University of Granada, Spain \\
    \textsuperscript{b} \href{https://lsi.ugr.es}{Department of Software Engineering (LSI)}, \href{https://dasci.es}{Andalusian Institute of Data Science and Computational Intelligence (DaSCI)}, University of Granada, Spain \\
    \vspace{3.5em}
  \end{center}
}

\begin{document}





\maketitle
\affiliations

\begin{abstract}

    Out-of-distribution (OOD) detection is a fundamental requirement for the reliable deployment of artificial intelligence applications in open-world environments. However, addressing the heterogeneous nature of OOD data, ranging from low-level corruption to semantic shifts, remains a complex challenge that single-stage detectors often fail to resolve. To address this issue, we propose SeNeDiF-OOD, a novel methodology based on Semantic Nested Dichotomy Fusion. This framework decomposes the detection task into a hierarchical structure of binary fusion nodes, where each layer is designed to integrate decision boundaries aligned with specific levels of semantic abstraction. To validate the proposed framework, we present a comprehensive case study using MonuMAI, a real-world architectural style recognition system exposed to an open environment. This application faces a diverse range of inputs, including non-monument images, unknown architectural styles, and adversarial attacks, making it an ideal testbed for our proposal. Through extensive experimental evaluation in this domain, results demonstrate that our hierarchical fusion methodology significantly outperforms traditional baselines, effectively filtering these diverse OOD categories while preserving in-distribution performance.

    
\end{abstract}


\section{Introduction}

Out-of-distribution (OOD) detection~\cite{yang2024generalized} has become a crucial process in the field of artificial intelligence, as many models working on high-risk applications require reliable input data to avoid producing unpredictable outputs that could have serious consequences.

OOD detection is required in a wide range of scenarios, spanning conventional settings with tabular data to more challenging contexts involving heterogeneous or multimodal inputs~\cite{liang2025diffusion}. Within computer vision, the ability to recognize when an image is outside the expected distribution can prevent critical failures. Computer vision models in many areas are subject to the open-world hypothesis~\cite{barcinablanco2023managing,zhu2024open}, with all the difficulties that this entails. OOD images are susceptible to being fed into models, either intentionally or casually, degrading their performance if not filtered correctly. Not all OOD images have the same nature or intentions, and each area of applicability is prone to different types of OOD inputs.

Several examples illustrate this concept. Medical diagnosis using images~\cite{zamzmi2024out} is generally a near closed-world application, as only medical images are expected during deployment. However, there remains a possibility of encountering images with unforeseen diseases that require early filtering. In contrast, autonomous driving systems~\cite{henriksson2023out,shoeb2025out} must be resilient to any unusual objects or unexpected road conditions not covered during training to ensure passenger safety. User applications employing vision models~\cite{pertusa2018mirbot,asani2023mpd,okyere2024deployment} must be capable of handling a wide range of scenarios. These applications may receive various OOD inputs, including unrelated images, whether accidental or as part of an adversarial attack, images relevant to the application but belonging to unrecognized classes, or images with inaccuracies, errors, or different semantic categories that still relate to the application's purpose. It is important to be able to detect all these kinds of OOD inputs and even to categorize them to perform an appropriate response to each of them.

Drawing a parallel with multi-class classification, where each class would represent either an in-distribution (in-distribution) input or a type of OOD input, we can consider nested dichotomies techniques~\cite{frank2004ensembles} to address this problem. Nested dichotomies break down classification tasks into multiple binary subproblems, each contrasting one category against the rest. During testing, each step determines whether the sample belongs to the split class. If it does not, the process continues to the next dichotomy. This sequence repeats until the dichotomy that assigns a class to the input is reached. Generally, nested dichotomies are less favored over other decomposition techniques like \emph{one-vs-one} or \emph{one-vs-all} \cite{knerr1990single} because, despite being more efficient, their performance depends significantly on the hierarchy of divisions within the dichotomies. However, when categories are already hierarchically structured, this limitation turns into an advantage.

Many computer vision deployments still operationalize OOD recognition as a single scalar score that separates in-distribution from OOD under an essentially unstructured notion of novelty. This abstraction is misaligned with open-world, user-facing pipelines, where OOD inputs arise from heterogeneous mechanisms and demand different responses, ranging from unrelated content to semantically adjacent but unsupported classes and acquisition degradations. Consequently, a monolithic score offers limited interpretability and provides weak guidance for system-level handling. A missing ingredient is a principled methodology that decomposes OOD recognition into semantically meaningful decisions, enabling robust filtering while characterizing distinct OOD modes.

A structured view of novelty can be obtained by organizing the input space into successive semantic checkpoints rather than treating it as a flat in-distribution versus OOD partition. In such a design, each checkpoint captures a coarse-to-fine distinction that mirrors how a practitioner would rule out implausible content before committing to a specific interpretation. This hierarchical organization encourages the use of specialized criteria at each stage and constrains error propagation, since ambiguous samples are intercepted early by broad semantic tests instead of being forced into a final decision. Beyond improving reliability, this methodology naturally yields a diagnostic outcome that indicates where and why an input becomes incompatible with the intended operating domain, which is valuable for monitoring, curation, and data acquisition strategies.

In this paper, we introduce SeNeDiF-OOD (Semantic Nested Dichotomies Fusion for OOD detection), a new methodology that casts OOD recognition as a structured fusion process over a semantic cascade of dichotomous decisions. In particular, SeNeDiF-OOD combines hierarchical decision-level fusion with semantic priors embedded in the nested dichotomies topology. These priors provide human-aligned guidance that facilitates OOD detection. At each level of the hierarchy, the method also integrates heterogeneous expert fusion, leveraging the representational capabilities of artificial intelligence (AI) models to better separate the underlying dichotomies. Overall, this design enables a principled fusion of human knowledge and machine intelligence, which can be beneficial for facilitating OOD detection, a setting in which many computer vision models still struggle~\cite{zhao2025neuralood}.

Beyond proposing an OOD score or a confidence metric, SeNeDiF-OOD provides a system-level framework for OOD handling based on a semantically grounded decomposition of the input space into a sequence of interpretable binary decisions, enabling the use of detection mechanisms where they are most effective. The reasons for proposing this methodology are as follows: (1) to achieve a more robust filtering of OOD images, as uncertain images must pass through multiple layers of dichotomies, each increasingly refined; (2) to leverage the decomposition provided by dichotomies, reducing the OOD detection problem in most layers to binary classification; and (3)  to utilize different layers to identify and categorize various types of OOD images. This last point also aims to bridge the gap between OOD detection and active learning systems \cite{xie2024deep,schmidt2025joint}, as the final layer of this system is designed to filter data that could define new, previously unknown classes within the problem being addressed.

For the experimental analysis of the proposed methodology, a monument style classification problem is used as a case study, leveraging the MonuMAI framework \cite{lamas2021monumai}. MonuMAI is a publicly available tool for architectural style analysis, implemented as a mobile application connected to multiple deep learning models for classification and detection of key architectural elements. Users capture photographs of monuments and submit them to the application, which returns the processed image with the detected elements localized and labeled, together with the predominant architectural styles. Overall, the nature of MonuMAI, together with the resources it provides in terms of curated data and deployed models, makes it a valuable testbed for exploring and benchmarking a broad range of artificial intelligence problems~\cite{diaz2022explainable}.

\begin{figure}[t!]
    \centering
    
    \begin{subfigure}[b]{0.24\textwidth}
        \centering
        \includegraphics[width=\textwidth]{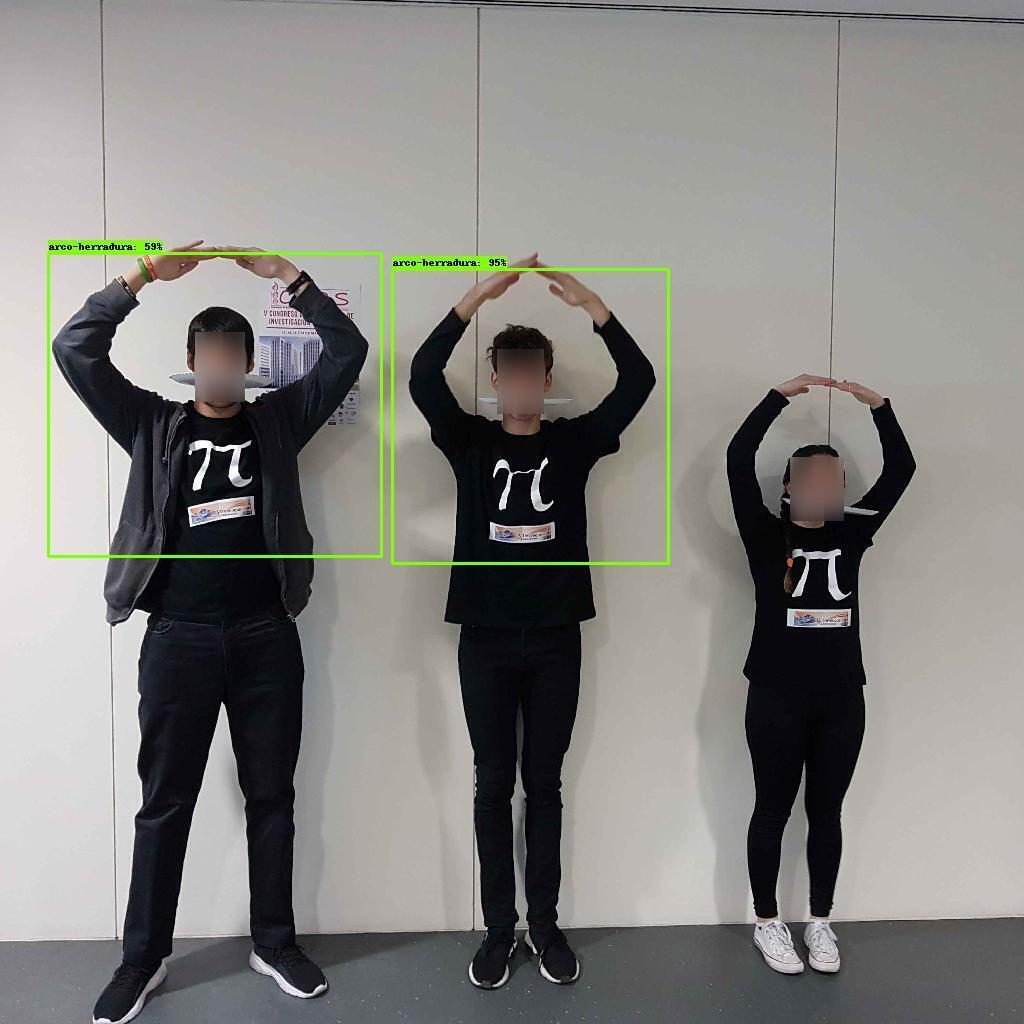}
        \label{fig:sub1}
    \end{subfigure}
    \hfill 
    \begin{subfigure}[b]{0.24\textwidth}
        \centering
        \includegraphics[width=\textwidth]{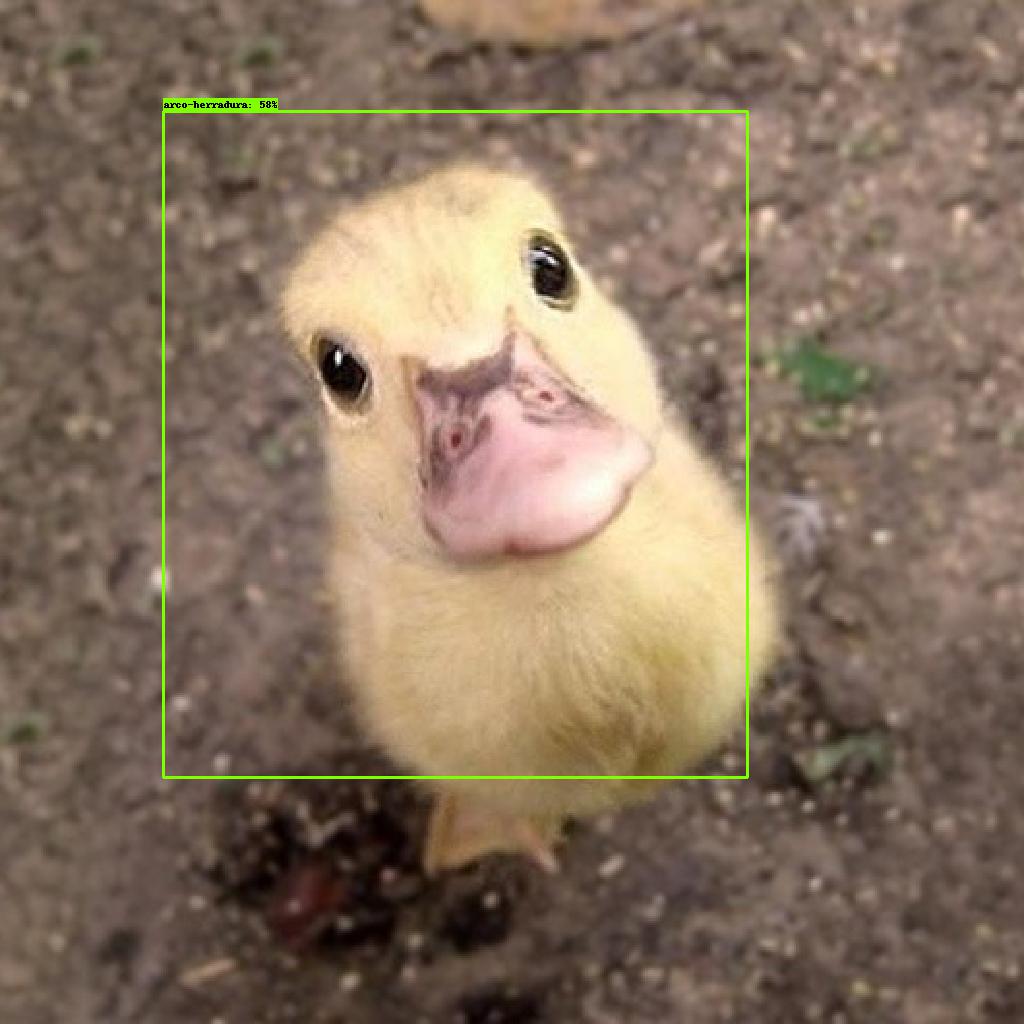}
        \label{fig:sub2}
    \end{subfigure}
    \hfill
    \begin{subfigure}[b]{0.24\textwidth}
        \centering
        \includegraphics[width=\textwidth]{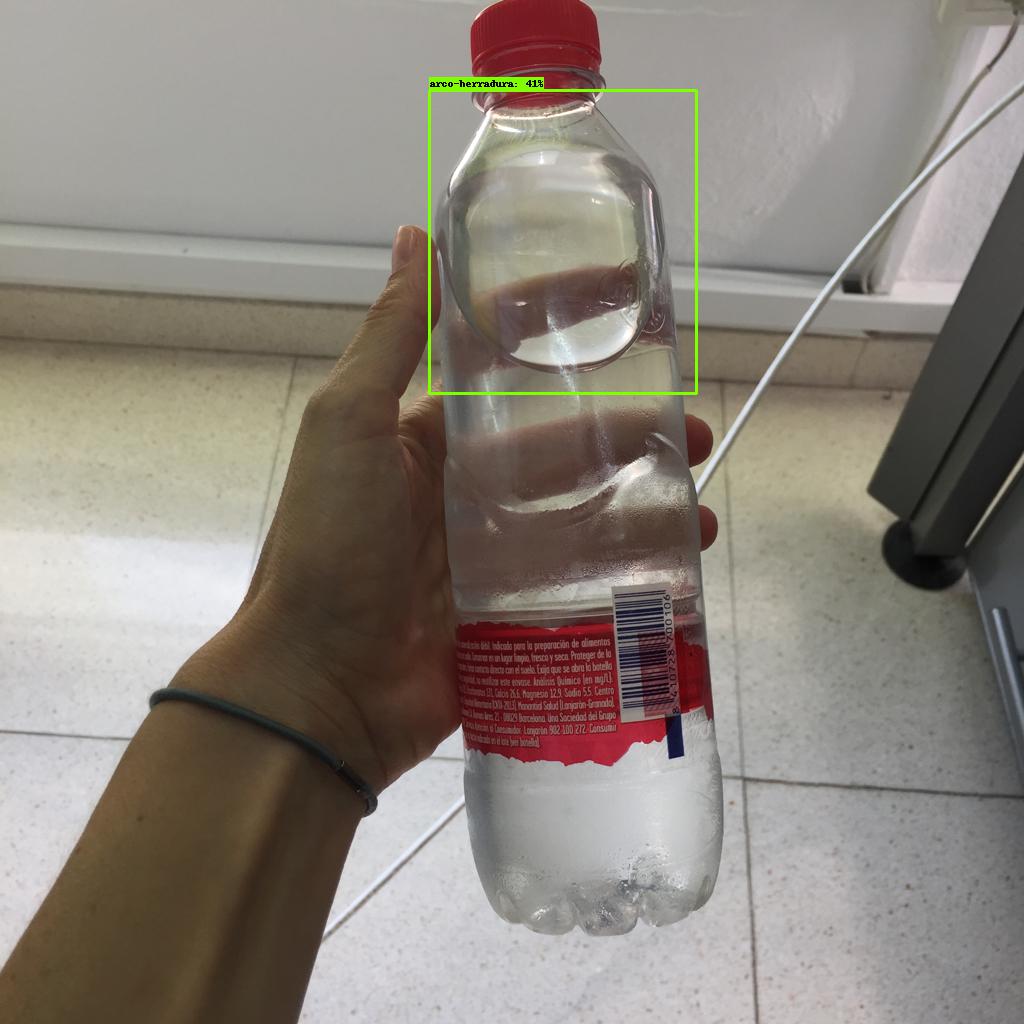}
        \label{fig:sub3}
    \end{subfigure}
    \hfill
    \begin{subfigure}[b]{0.24\textwidth}
        \centering
        \includegraphics[width=\textwidth]{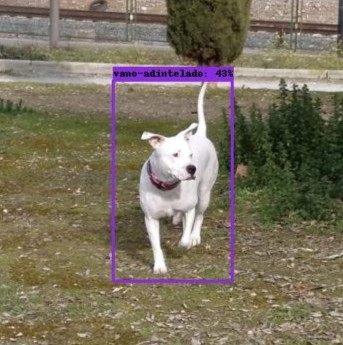}
        \label{fig:sub4}
    \end{subfigure}
    
    \caption{Examples of MonuMAI misclassifying OOD inputs as architectural elements.
    }
    \label{fig:OOD_examples}
\end{figure}

In an open, user-facing setting, MonuMAI is naturally exposed to diverse inputs that extend beyond its intended scope. Over more than four years of operation, the server has received thousands of images, including non-monument content, low-quality captures (blurred, rotated, cropped), buildings with limited architectural relevance, and monuments exhibiting styles outside the supported set. Moreover, some users deliberately submit misleading content, exploiting reflections or repetitive patterns that trigger spurious detections. Figure \ref{fig:OOD_examples} shows several examples where the MonuMAI application fails to detect OOD user inputs. Since the underlying model is trained on four architectural styles (hispanic-muslim, gothic, renaissance, and baroque), inputs from unseen styles also often induce confident misclassifications with plausible but incorrect elements and labels. These situations exemplify OOD challenges in which user inputs deviate substantially from the training distribution, motivating the need for a methodology that can properly filter and characterize OOD images.

Using a dataset composed of images uploaded by MonuMAI users, complemented with additional images obtained from public datasets covering different domains, we conducted an experiment to compare SeNeDiF-OOD methodology against the MonuMAI deep learning model itself. The results confirm that SeNeDiF-OOD significantly reduces false detections of monuments compared to the original model while preserving in-distribution images. Additionally, we analyze the system layer by layer, evaluating the effectiveness of each layer individually and drawing conclusions about the types of OOD images filtered at each stage.

potential of nested dichotomies to OOD detection was initially explored in \cite{antequera2024establishing}, where a preliminary filtering layer was tested on MonuMAI. This paper goes far beyond that initial exploration, introducing a completely original and in-depth development of the approach. We not only refine and formalize the methodology but also provide a rigorous theoretical and experimental analysis that significantly improves the OOD robustness of the original MonuMAI model. At the theoretical level, we  formulate a generalizable methodology for hierarchical decision fusion in image-based problems, offering a precise characterization of the different types of OOD that can be identified. On the experimental side, we instantiate and validate this architecture within a real-world pipeline, conducting a thorough evaluation in the context of architectural style and element detection, leveraging the MonuMAI framework.


The structure of this paper is as follows. Section \ref{sec:background} provides the necessary background on out-of-distribution detection and nested dichotomies. Section \ref{sec:nested_theory} establishes the theoretical foundations of the proposed SeNeDiF-OOD methodology, formulating a generalizable methodology for hierarchical decision fusion. Section \ref{sec:nested_monumai} details the practical implementation of the SeNeDiF-OOD methodology within the MonuMAI framework, outlining the established layers and the training procedures for each. Section \ref{sec:experiments} presents the experiments conducted on this dichotomy-based system, comparing its performance with the original MonuMAI version and analyzing the results both globally and at the layer level. Finally, Section \ref{sec:conclusions} concludes with a summary of the work and key takeaways.

\section{Background} \label{sec:background}

This section introduces the fundamental concepts underpinning the methodology proposed in this study: out-of-distribution detection and nested dichotomies.

\subsection{Out-of-distribution detection}

Out-of-distribution detection has emerged as a fundamental component for ensuring the robustness, safety, and trustworthiness of modern machine learning systems~\cite{yang2024generalized,hendrycks2021unsolved}. In real-world deployment, models frequently encounter inputs that deviate significantly from the statistical distribution of their training data. These inputs, referred to as OOD samples, can originate from different semantic categories, domains, acquisition conditions, or even adversarial manipulations. If not properly identified, OOD samples may be misclassified with high confidence, potentially leading to severe consequences in high-stakes domains such as autonomous driving, healthcare, or financial systems.

The importance of OOD detection lies in its role as a safeguard mechanism. For example, an autonomous driving system trained primarily on urban street scenes should raise an alert when faced with unexpected rural or adverse weather conditions, instead of producing overconfident predictions that may compromise safety~\cite{henriksson2023evaluation,nitsch2021out}. Similarly, in medical imaging, diagnostic systems must detect when a sample corresponds to a rare disease or a novel variant not represented in training data, in order to avoid erroneous diagnoses~\cite{karimi2022improving,wollek2024out,barandas2024evaluation}. Likewise, in natural language processing, the rise of foundation models and their applicability as autonomous agents makes it necessary to further develop robust techniques for handling text inputs that fall outside the scope anticipated during system design~\cite{huang2025atcaf,yuan2023revisiting}. The primary goals of OOD detection are therefore to improve system reliability by preventing harmful misclassifications, to facilitate safer deployment of machine learning models, and to enable iterative model refinement by flagging unexpected cases for human inspection~\cite{azizmalayeri2022your}.

Despite its intuitive formulation, OOD detection is a challenging problem due to the complex and dynamic nature of real-world data. Distributional shifts can be broadly categorized into two main types: covariate shift and semantic shift. Covariate shift occurs when the marginal input distribution changes while the label space remains constant, such as when style, lighting, or domain characteristics differ between training and test data. In contrast, semantic shift implies changes in the label distribution, for instance through the introduction of entirely new categories, and is considered the primary focus of most OOD detection research. While covariate shifts primarily affect generalization performance, semantic shifts demand explicit mechanisms for rejection, since models should abstain from making predictions for categories outside the training label space.

Over the past few years, a wide range of methodological approaches have been developed to tackle OOD detection. These can be broadly grouped into classification-based, density-based, distance-based, and reconstruction-based methods. Classification-based approaches leverage the outputs or internal representations of neural networks, with techniques such as maximum softmax probability, energy-based scoring, or activation rectification~\cite{hendrycks2016baseline,liu2023gen}. Density-based methods estimate the probability distribution of in-distribution data and flag samples with low likelihood as OOD, although generative models often struggle with high-dimensional data~\cite{hendrycks2019scaling,liu2020energy}. Distance-based approaches rely on computing feature-space distances to class prototypes or centroids, with Mahalanobis distance and nearest-neighbor methods being notable examples~\cite{lee2018simple,sun2022out,sevillano2025stood}. Reconstruction-based methods, in turn, exploit differences in reconstruction error between in-distribution and OOD samples using autoencoders or generative models~\cite{denouden2018improving}. Each family of methods presents strengths and limitations, and recent advances have combined these perspectives to improve robustness.

\subsection{Nested dichotomies for multi-class problems}

Nested dichotomies~\cite{frank2004ensembles} constitute a structured decomposition framework for tackling multi-class classification problems by reducing them into a hierarchy of binary decisions. Unlike traditional strategies such as one-vs-one or one-vs-all, nested dichotomies organize class discrimination into a binary tree structure, where each internal node corresponds to a binary classifier that separates a subset of classes into two disjoint partitions. This recursive partitioning continues until each node corresponds to a single class.

This hierarchical formulation has several advantages. First, it enables the transformation of any base binary classifier into a multi-class classifier, making it a versatile learning approach. Second, by breaking down the original problem into smaller, more tractable subproblems, it can potentially exploit inductive biases of base learners more effectively. Each binary classifier in the tree only needs to distinguish between two subsets of classes, which can be particularly beneficial when class boundaries are complex or when classes exhibit strong similarity within subsets.

Formally, let $C$ be the original set of $k$ classes. A nested dichotomy defines a binary tree with $k-1$ internal nodes. At each node, the current set of classes is partitioned into two non-empty, non-overlapping subsets $C_L$ and $C_R$. A binary classifier is trained to distinguish between instances belonging to $C_L$ and $C_R$. Then, the same procedure is recursively applied to both $C_L$ and $C_R$ until all subsets are singleton sets. The training of the nested dichotomy requires $k-1$ classifiers, each trained on a subset of the data where the target labels are coarsened to the current binary split.

The hierarchical nature of nested dichotomies also introduces variance due to the many possible class partitionings. For $k$ classes, there are $(2k-3)!!$ possible binary trees\footnote{The operator $!!$ denotes the \emph{double factorial}, defined as $m!! = \prod_{i=0}^{\lceil m/2  \rceil - 1} (m-2i) = m (m-2)(m-4)\dots$}, each corresponding to a different decomposition strategy. This variability can significantly impact performance. Several tree selection strategies have been proposed, such as random nested dichotomies~\cite{frank2004ensembles}, balanced nested dichotomies~\cite{dong2005ensembles} or performance-based selection~\cite{leathart2019ensembles}.

\section{The SeNeDiF-OOD Methodology for Image Classification} \label{sec:nested_theory}


This section presents the SeNeDiF-OOD methodology to approach OOD image detection through the lens of nested dichotomies. Rather than relying on a single global rejection rule, the approach operationalizes OOD handling as a sequence of semantically grounded binary decisions, where each stage can adopt the most appropriate detection mechanism for its local data regime. In contrast to open-set recognition setups that typically presuppose a narrow semantic domain, the method captures progressively broader OOD categories, from fully unrelated inputs to near-distribution but unseen classes. Moreover, while hierarchical decision processes are often designed for label assignment within a closed taxonomy, SeNeDiF-OOD integrates explicit rejection at each stage, enabling principled handling of open-world uncertainty instead of forced classification.

Section~\ref{ssec:ood_categories} introduces a taxonomy of OOD image types, illustrating the variety of OOD scenarios that can arise across different applications. Section~\ref{ssec:nested_dicotomies_ood} formalizes the SeNeDiF-OOD architecture, constructing a semantically grounded hierarchy that maps specific OOD categories to distinct decision fusion nodes. Lastly, Section~\ref{ssec:nested_responsible} situates the proposed methodology within the broader context of responsible artificial intelligence, emphasizing key considerations such as robustness, explainability, and human alignment.




\subsection{Open-world image classification: identifying OOD categories}~\label{ssec:ood_categories}

OOD inputs are typically defined as data samples that differ significantly from the training distribution of a model. However, this seemingly straightforward concept becomes far more nuanced when we consider its manifestations in real-world computer vision settings. There is not just one kind of OOD image. Instead, we are confronted with a diverse and unpredictable landscape of OOD cases, each capable of derailing even the most carefully trained models.

Several types of OOD inputs warrant caution to prevent vision models from producing inappropriate outputs:
\begin{itemize}
    \item \textbf{Malicious OOD inputs.} They are often stem from adversarial attacks or other exploits that leverage the model weaknesses. These images are not only out-of-distribution but intentionally crafted to deceive the model, exploiting weaknesses in the decision boundaries of the models. Unlike accidental OOD inputs, these are targeted and present an elevated risk in high-stakes applications.

    \item \textbf{Accidental OOD inputs.} These inputs might be provided in error. They can be either similar or dissimilar to the context of the problem. Dissimilar-context images may have been provided accidentally during submission, while similar-context image submissions may result from a partial misunderstanding of the operational design domain of the model. While unintentional, these mistakes highlight the need for models that can gracefully reject irrelevant data instead of confidently misclassifying them.

    \item \textbf{Degraded images.} These include inputs with heavy noise, extreme lighting conditions, motion blur or compression artifacts. Although these images may still be of scope in terms of subject matter, the training distribution of the models is not likely to cover these distortions comprehensively.

    \item \textbf{Novel classes.} These images are contextually relevant to the task but belong to semantic categories the model was never trained on. For instance, a model trained to classify common dog breeds might encounter an unfamiliar breed that has never seen during training. These cases often blur the line between OOD detection and open-set recognition, further complicating the problem. The novel classes can subsequently be used to retrain the model, enabling their recognition through an active learning process.
\end{itemize}

In sum, the spectrum of OOD images is broad, multi-faceted, and highly application-dependent. Effective solutions must account for this heterogeneity. The following section demonstrates how nested dichotomies can contribute to the filtering of different types of OOD inputs.

\subsection{SeNeDiF-OOD World Hierarchy and Layers}~\label{ssec:nested_dicotomies_ood}

Building on the concept of nested dichotomies, we can apply SeNeDiF-OOD to detect OOD samples in image classification. Nested dichotomies involve filtering inputs through multiple layers, each designed to handle different aspects of OODness, progressively narrowing down the scope of relevant inputs. In the context of image OOD detection, this method enables the system to distinguish between various types of OOD inputs by classifying them through a series of semantic decision boundaries. Each layer filters out inputs that either fall outside the scope of the problem or do not conform to the characteristics expected by the model, ultimately allowing for a more refined and accurate classification process through the fusion of decisions across layers, where each layer is handled by a dedicated expert detector trained for a specific purpose.

We now theoretically analyze the \emph{worlds} that should be present in an OOD semantic nested dichotomies system following the SeNeDiF-OOD methodology and outline their key aspects. Each layer will then consist of a dichotomy between two consecutive worlds, starting with the most distant worlds in the first layer and progressing to the nearest worlds in the final layer. We expect these guidelines to be applicable to any computer vision OOD detection problem, with each world enabling a semantically grounded fusion of hierarchical meaning that is adaptable to the specific context of the problem. 

\begin{itemize}
    \item \textbf{The far-world: completely unrelated inputs.} The first world, which we have denoted the \emph{far-world}, contains images that are completely unrelated to the problem domain. These images could belong to entirely different problem spaces, such as medical imaging or datasets like MNIST, when the task involves something completely different, such as dog breed classification. The \emph{zero-layer} of the system will consist in filtering out inputs from the far-world. It is crucial because it quickly rejects inputs that are not even remotely relevant to the task, ensuring that the model is not burdened with data that could cause it to fail outright. It can usually be addressed effectively through far-OOD detectors~\cite{winkens2020contrastive}. The remaining inputs will belong to the \emph{near-world}, described in the next paragraph. The far-world is home to most types of OOD inputs, highlighting accidentally submitted images and certain types of attacks.

    \item \textbf{The near-world: unrelated content in a similar feature space.} Next, the \emph{near-world} consists of images that share similar feature spaces with the problem images but are not part of the same task. These images typically follow the same general characteristics (e.g., they may be real-world colored photographs) but contain content that does not fit into the problem context. For example, in dog breed classification, images of people, vehicles or buildings might share the same visual characteristics—real-world photographs with similar lighting and color patterns—but are not relevant to the task at hand. Typically, near-world OOD identifications can be addressed through near-OOD detectors~\cite{fort2021exploring}, though since the near-world boundary is already defined by the zero-layer, filtering this world can be performed through a binary classification between the near-world and the next world (which is defined in the next paragraph). The near-world also hosts accidentally submitted images and more types of attacks, since the similarity to the problem distribution increases.

    \item \textbf{The superclass hierarchy: gradually narrowing down context.} Following the near-world, we introduce the \emph{superclass hierarchy}. This is a set of progressive worlds, starting from representations of high-level classes, narrowing down to more specific subclasses that relate to the problem. For each pair of consecutive worlds in this hierarchy, a new decision layer of the dichotomies system categorizes the images into increasingly fine-grained categories that still belong to the broader problem space. For instance, in the dog breed classification example, the first world in the class hierarchy could represent any animal. The worlds would then progressively follow the taxonomic ranks, highlighting several points of interest, such as mammals or canines. At these worlds, we begin to expect OOD accidental inputs due to misunderstandings of the problem, while still anticipating various types of attacks as the relationship to the real domain increases.

    \item \textbf{The problem full world: expanding beyond known classes.} At the next level, we have the \emph{problem full world}. This world acknowledges that there may be classes related to the problem domain that the model has not been trained on. For example, in the dog breed classification task, the model may have been trained on 50 specific breeds, but the full-problem world encompasses all dog breeds, even those not represented in the training data. A penultimate decision layer in the dichotomies system can perform binary classification between the nearest superclass images and the problem full world images. This layer leverages the generalization capability of such a classifier, enabling it to set a boundary on the problem full world by using nearly as much training data as those used for the true distribution. For instance, the breeds used to train the model, after minimal refinement to establish a clearer boundary with the nearest superclass, could define the \emph{dog} class in the nested dichotomy, as opposed to the \emph{canines} nearest superclass. In the full problem world, we primarily expect novel classes as OOD inputs, which can later be used to retrain a more comprehensive model once enough samples are filtered in the last layer described below.

    \item \textbf{The known-world: the exact training distribution.} Finally, the \emph{known-world} represents the precise training distribution on which the model was trained on. The \emph{last layer} has the goal of filtering out the novel classes that are not represented in the training set, but are considered to belong to the full-problem world. For the
    last layer, it will be usually necessary to resort to a specialized OOD detector, since it may not be feasible to have a full distribution of the problem unknown world. Detectors with outlier exposure~\cite{hendrycksdeep} may help to define a clearer border between known and unknown worlds and facilitate novelty detection. This world represents the comfort zone of the model, where it can make confident predictions based on its learned parameters. This world is the in-distribution world: no OOD inputs are expected in this world.
\end{itemize}

This hierarchical structure of OOD categories can be naturally addressed using nested dichotomies, by directly extending the approach typically employed for multi-class problems. In this context, each OOD category can be treated as a distinct class, after excluding its internal categories to prevent overlap. Furthermore, the hierarchical organization of the categories enables the construction of a canonical and semantically meaningful dichotomy tree. At each node, the objective is to distinguish the outermost category from the remaining ones, progressively refining the classification until reaching the in-distribution category in the final split. This design avoids reliance on a randomly generated dichotomy tree, providing a more structured and interpretable decision process.

The underlying decomposition performed using the SeNeDiF-OOD methodology is domain-agnostic and can be transferred to other image-based applications. In general, the procedure consists of: (i) defining a far-world corresponding to inputs that are visually incompatible with the target domain; (ii) specifying a near-world that captures the broad data modality (e.g., natural images); (iii) introducing one or more semantic superclasses that progressively narrow the context of interest; (iv) defining a problem full world that includes both known and unknown task-relevant classes; and (v) isolating the known world corresponding to the training distribution.

\begin{figure}[htbp]
    \centering
    \includegraphics[width=\textwidth]{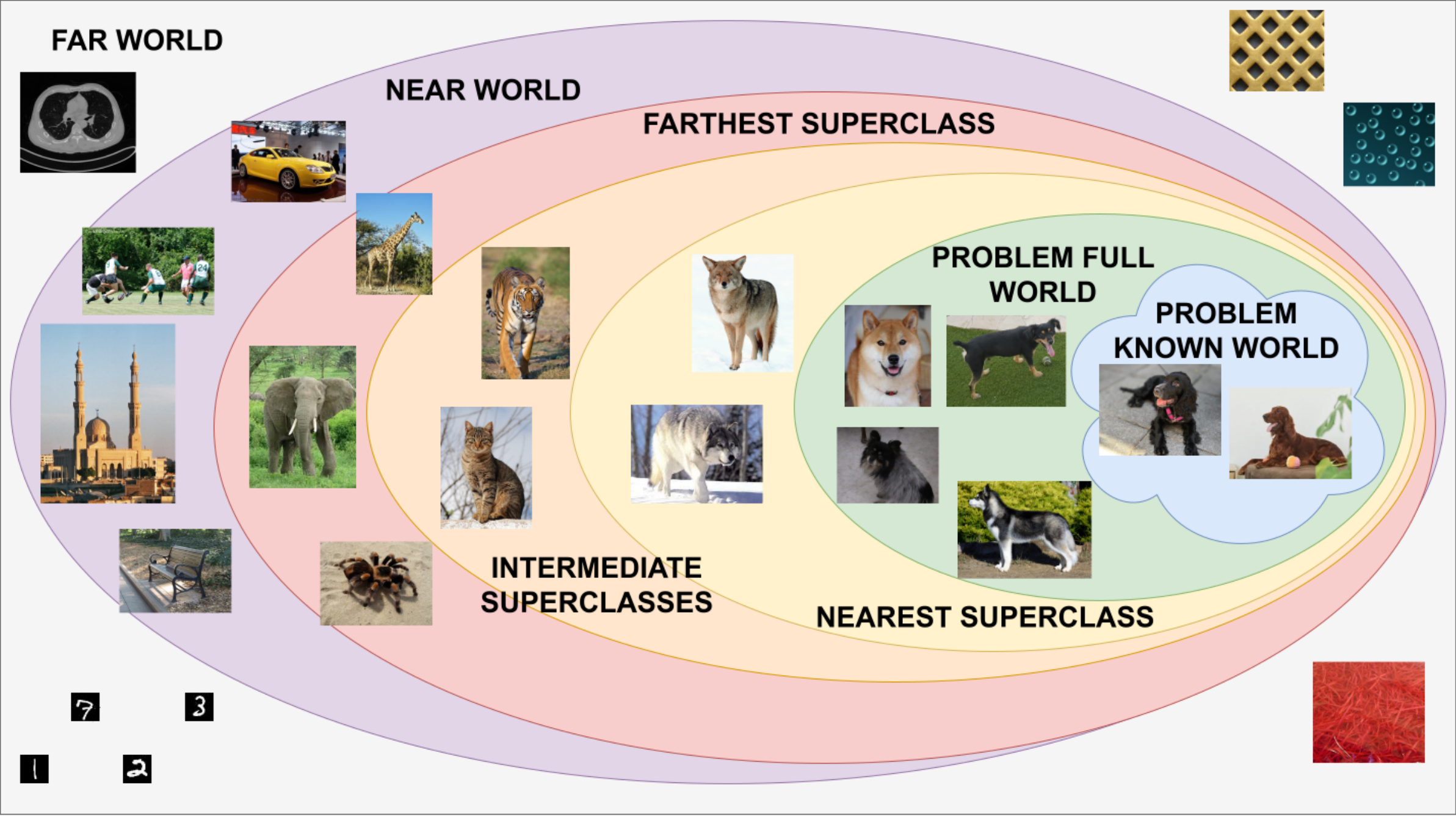}
    \caption{Nested set view of the data regime: the far world surrounds the near world,
followed by progressively closer superclasses until reaching the problem’s known world. The
figure motivates layered OOD handling: coarse filters are effective at the periphery, whereas
near the known world, decisions must be semantics-aware and carefully calibrated.}
    \label{fig:nested_class_hierarchy}
\end{figure}
\begin{figure}[htbp]
    \centering
    \includegraphics[width=\textwidth]{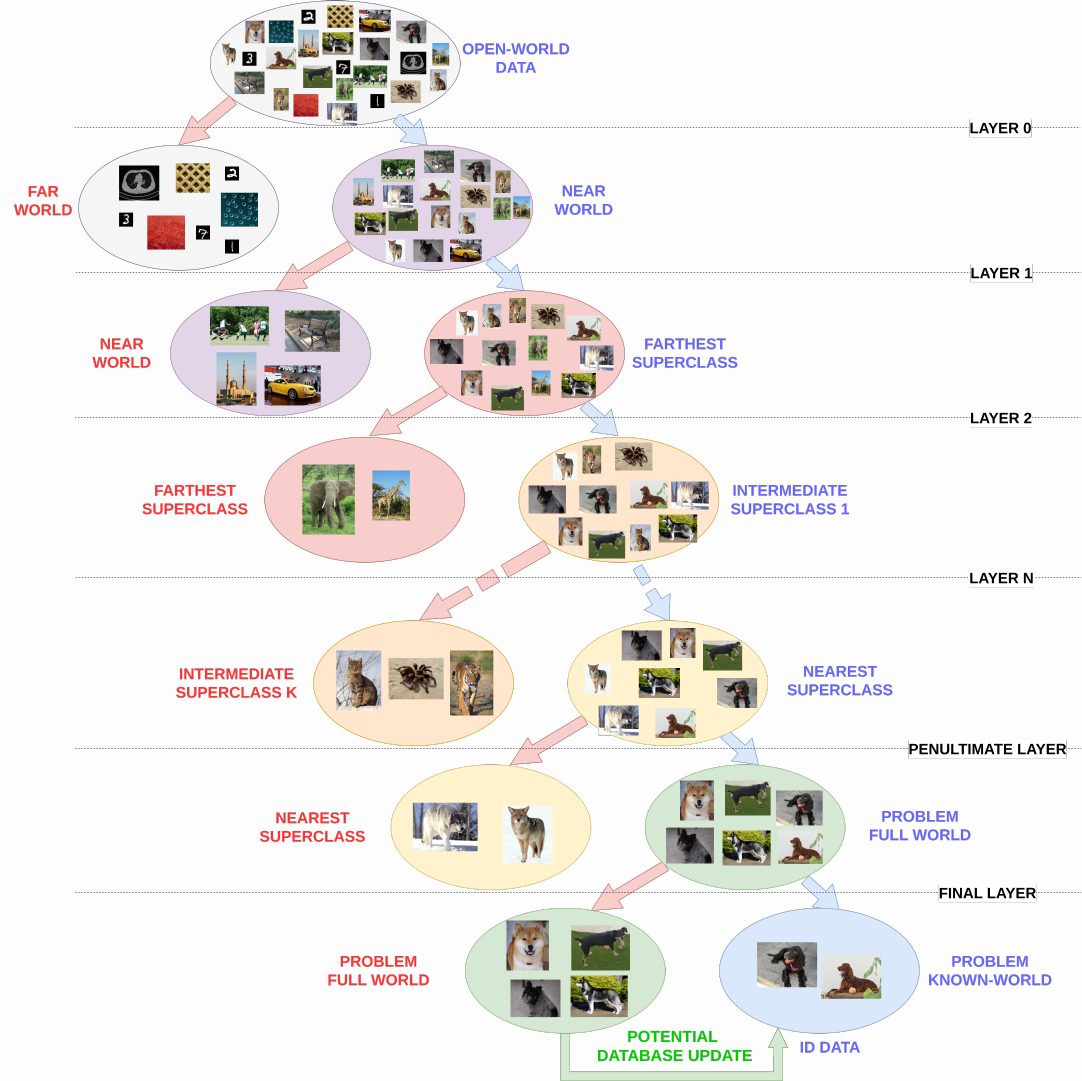}
    \caption{Schematic of the proposed cascade with nested dichotomies. Each layer enforces a
semantically aligned gate. Samples flow top-down: at every stage, out-of-scope inputs are
rejected, and the remainder becomes progressively more specific, from open-world data to the
task’s in-distribution data.}
    \label{fig:nested_class_dichotomies}
\end{figure}

Figures~\ref{fig:nested_class_hierarchy} and~\ref{fig:nested_class_dichotomies} show, respectively, the worlds and layers of the nested dichotomies system obtained by applying the SeNeDiF-OOD methodology to the dog breed example described above. In Table~\ref{tbl:nested_worlds} we summarize the description of the worlds, the types of expected OOD inputs and provide more examples.

\begin{table*}[t]
\centering
\caption{Taxonomy of worlds for OOD analysis with descriptive and example-based views.}
\label{tbl:nested_worlds}

\emph{General description}\\[0.5ex]
\begin{tabular}{p{0.18\textwidth} p{0.44\textwidth} p{0.33\textwidth}}
\toprule
\textbf{World} & \textbf{Description} & \textbf{Expected OOD} \\
\midrule
Far-world        & Completely unrelated images. & Attacks (+), Accidental mistakes (++) \\
\addlinespace[0.6ex]
Near-world       & Unrelated context, same type of images. & Attacks (+++), Accidental mistakes (++) \\
\addlinespace[0.6ex]
Superclass hierarchy & Progressively fine-grained superclasses bounding the context of the problem. & Attacks ($\downarrow$), Misunderstanding mistakes ($\uparrow$) \\
\addlinespace[0.6ex]
Problem full world     & The full context of the problem, considering known and unknown classes. & Novel classes (+++) \\
\addlinespace[0.6ex]
Known world      & The in-distribution set. & None \\
\bottomrule
\end{tabular}

\vspace{1em} 

\emph{Illustrative examples}\\[0.5ex]
\begin{tabular}{p{0.17\textwidth} p{0.25\textwidth} p{0.25\textwidth} p{0.25\textwidth}}
\toprule
\textbf{World} & \shortstack{\textbf{Example 1}\\ \footnotesize (Dog breed classification)} & \shortstack{\textbf{Example 2}\\ \footnotesize (Handwritten digit recognition)} & \shortstack{\textbf{Example 3}\\ \footnotesize (Monument style classification)} \\
\midrule
Far-world        & \textit{EMNIST}, \textit{Textures}, medical images, ... & \textit{ImageNet} & \textit{EMNIST}, \textit{Textures}, medical images, ... \\
\addlinespace[1ex]
Near-world       & \textit{ImageNet} & Document analysis & \textit{ImageNet} \\
\addlinespace[1ex]
Superclass hierarchy & Animals $\rightarrow$ Mammals $\rightarrow$ Canines & Text recognition $\rightarrow$ ... $\rightarrow$ Handwritten recognition & Buildings \\
\addlinespace[1ex]
Problem full world     & Dogs & Handwritten digits and letters (\textit{EMNIST}) & Monuments \\
\addlinespace[1ex]
Known world      & Known breeds & Handwritten digits (\textit{MNIST}) & Known architectural styles \\
\bottomrule
\end{tabular}

\vspace{0.6ex}
    \footnotesize
    \emph{Notes.} Plus signs indicate relative prevalence or severity of the phenomenon: $(+)$ low, $(++)$ medium, $(+++)$ high. Arrows indicate expected trend when moving down the hierarchy: $(\uparrow)$ increase, $(\downarrow)$ decrease.

\end{table*}

\subsection{SeNeDiF-OOD for a responsible AI}~\label{ssec:nested_responsible}

In this section we discuss what SeNeDiF-OOD methodology can contribute to OOD detection from the paradigm of AI safety and reliance. We distinguish three fronts of interest: robustness, explainability and human alignment.

\begin{itemize}
    \item \textbf{SeNeDiF-OOD} and robustness. Robustness in AI models refers to their capacity to maintain performance when exposed to perturbations or adversarial inputs~\cite{tocchetti2025ai}. In a system based on nested dichotomies, the multiple layers-each specialized in a distinct form of filtering-form an ensemble that enhances the detection of various types of anomalies. Adversarial examples, in particular, must pass through a sequence of filters, each trained with different criteria, which increases the likelihood of their rejection before reaching the final classifier. Other forms of perturbation are similarly likely to be prevented to belong to the inner worlds of the system.

    \item \textbf{SeNeDiF-OOD} and explainability. Explainability is essential in many AI applications, as it enables understanding of the rationale behind the decisions of an algorithm~\cite{ali2023explainable, ullah2024explainable}. Explainability can also be fundamental in OOD detection, as it provides insights into why a given input has been classified as in-distribution or rejected as out-of-distribution~\cite{choi2023concept,martinez2023novel}. In the context of multi-class classification, the tree-like structure of nested dichotomies offers a natural framework for model interpretability~\cite{fdez2021learning}. Because decisions are made sequentially at each node, it becomes possible to trace the decision path and identify the specific nodes responsible for a given outcome. This facilitates both the interpretation of predictions and the diagnosis of potential errors. This interpretability extends naturally to the use of nested dichotomies for OOD detection. By examining the specific layer at which an input is filtered out, we can infer the nature of the input and the reason for its rejection. Moreover, if an input is incorrectly filtered, the responsible node can be easily identified, allowing for targeted corrections or retraining.

    \item \textbf{SeNeDiF-OOD} and human alignment. AI alignment \cite{fel2022harmonizing,muttenthalerhuman2023} has gained momentum in recent years due to the increasing need for artificial intelligence models to make decisions that are consistent with human objectives. Muttenthaler et al. \cite{muttenthaler2024aligning} highlight that, in vision models, a key misalignment often arises because these models fail to capture the various levels of abstraction present in human knowledge. Typically, human knowledge can be organized hierarchically through different levels of grouping. Muttenthaler et al. further demonstrate that incorporating such structured knowledge representations into learning models enhances their performance and allows for a closer approximation of human behavior and uncertainty. The SeNeDiF-OOD methodology follows this general principle by learning progressively fine-grained hierarchical representations of knowledge. This facilitates decision-making in OOD detection that is more aligned with human reasoning.
\end{itemize}

\section{SeNeDiF-OOD and MonuMAI: a case study}\label{sec:nested_monumai}



This section presents the application of the SeNeDiF-OOD methodology within the MonuMAI style classification pipeline.  Given its open nature, the system is inherently exposed to the risk of receiving images that fall outside the intended scope of the application. Thus, MonuMAI serves as a practical use case for the proposed methodology. First, we present the complete MonuMAI framework and describe its main features. Then, a suitable class hierarchy is defined to represent the OOD categories relevant to MonuMAI, enabling the construction of the semantic hierarchical structure where SeNeDiF-OOD methodology will be applied. Subsequently, the implementation of each filtering layer within this hierarchical framework is detailed.

\subsection{The MonuMAI Monument Detection API}

The MonuMAI (Monument with Mathematics and Artificial Intelligence) framework~\cite{lamas2021monumai} constitutes a comprehensive initiative that combines computer vision, expert knowledge in art history, and citizen science to facilitate the automatic recognition of architectural styles and elements in monumental heritage. The primary goal of MonuMAI is to democratize access to cultural heritage understanding by leveraging deep learning tools to emulate the expertise of architectural historians through image analysis.

MonuMAI was developed in response to the challenges involved in analyzing monuments, such as the overlap of architectural elements across different styles, variations in the preservation state of buildings, and the presence of noise or distortions in the images. Prior methods \cite{hesham2021monuments,llamas2017classification,shalunts2012classification,zhang2014recognizing,zhao2018architectural} exhibit limitations, including scalability across styles or elements, susceptibility to image perspective distortions, and restricted applicability under uncontrolled real‑world conditions.

The framework was first introduced at the European Researchers' Night 2018 in Granada, emphasizing its relevance and potential to foster public engagement with art history. The initiative represents a collaborative effort between the University of Granada (UGR) and the Fundación Descubre, reflecting both a strong academic basis and a firm commitment to community involvement.

At the core of MonuMAI lies a novel taxonomy specifically devised to capture the relationships between characteristic architectural elements and four major European styles spanning from the 8th to the 18th century: Hispanic-Muslim, Gothic, Renaissance, and Baroque. This hierarchical taxonomy was constructed in close collaboration with domain experts and is structured as a tree with style categories at the first level and their associated key elements at subsequent levels. It reflects the historical continuity and stylistic overlaps between these architectural movements, providing a robust semantic foundation for automatic classification and detection tasks.

To implement this taxonomy, the MonuMAI dataset was created as a multi-task resource annotated with expert knowledge. It includes 1514 high-resolution images of monument facades, each labeled with its dominant architectural style. Additionally, 6650 instances of fifteen distinct architectural elements (e.g., horseshoe arch, pointed arch, lintelled doorway, solomonic column) were annotated with bounding boxes. The dataset exhibits a natural imbalance in element frequencies, which introduces realistic challenges for detection tasks and reflects their distribution in the built heritage. This dual annotation enables both classification and object detection pipelines and opens avenues for explainability analysis.

The MonuMAI deep learning pipeline comprises two main components. The first is MonuNet, a custom lightweight convolutional neural network~\cite{wiatowski2017mathematical} designed for architectural style classification. MonuNet achieves high accuracy while maintaining computational efficiency, making it suitable for deployment on mobile devices. The second component is MonuMAI-KED, a detection model built upon the Faster R-CNN~\cite{ren2015faster} framework with a ResNet-101 backbone~\cite{he2016deep}. This model locates and classifies the architectural elements in input images, offering fine-grained, interpretable information about stylistic features.

To bring this technological framework into real-world usage, the MonuMAI mobile and web application was developed as a citizen science platform. The app allows users to capture or upload images of monuments, which are then processed via the MonuMAI pipeline. The system returns annotated visualizations indicating the recognized style and key architectural elements. This participatory model not only disseminates cultural knowledge to the public but also enables the crowdsourced expansion of the MonuMAI dataset, as newly submitted images can be reviewed and incorporated by experts.

Experimental evaluations demonstrate the effectiveness of the proposed models under real-world conditions. 
The MonuMAI application is accessible on both iOS and Android, enabling users to actively engage in this citizen science initiative. To date, it has attracted more than 3,500 participants and gathered over 11,700 photographs, thereby contributing to an expanding dataset that supports ongoing research and model refinement. Complementary resources and detailed information about the project are available on the MonuMAI website\footnote{\url{https://monumai.ugr.es/}}.

\subsection{Semantic OOD Categories for MonuMAI}

Following the SeNeDiF-OOD methodology described in Section~\ref{ssec:nested_dicotomies_ood}, we aim to align each of the OOD worlds or categories described therein with categories that are meaningful for the MonuMAI OOD hierarchy. To this end, the following mappings are established:

\begin{itemize}
    \item \textbf{MonuMAI far-world: } The proper far-world (excluding the inner OOD categories) can be regarded as the complement of the near-world, which will be described in the following paragraph. Given the nature of monument photographs, the near-world can be clearly defined, with any input that does not meet these conditions being assigned to the far-world.

    \item \textbf{MonuMAI near-world: } Since the MonuMAI domain consists of color photographs of monuments, this area can be generalized to define the near-world of MonuMAI. Accordingly, we propose considering the near-world as the domain of color images representing real-world scenarios. A dataset such as ImageNet~\cite{deng2009imagenet} can be leveraged to define the near-world domain for MonuMAI.

    \item \textbf{Buildings: } Focusing now on the superclass hierarchy, in the case of MonuMAI we identify a single but fundamental superclass: buildings. This category provides a clear boundary for the open-world domain, as it subsumes the specific domain of MonuMAI while encompassing many other categories unrelated to monuments. It constitutes the first filtering stage in which the target domain begins to take shape, progressively refined through the application of nested dichotomies.

    \item \textbf{Monuments: } The problem full-world is defined as the complete domain of monuments, encompassing monuments from all existing architectural styles. It is essential to achieve a comprehensive representation of these styles, extending beyond those included in the MonuMAI domain, in order to establish a meaningful decision boundary with respect to the broader category of buildings.

    \item \textbf{Monuments with known-styles: } Finally, the known world or in-distribution set comprises the monuments belonging to the architectural styles used during the training of MonuMAI. This domain is explicitly defined by the training dataset employed in the development of the MonuMAI model.

\end{itemize}

\subsection{SeNeDiF-OOD layers for MonuMAI}

\begin{figure}[ht]
    \centering
    \includegraphics[width=\linewidth]{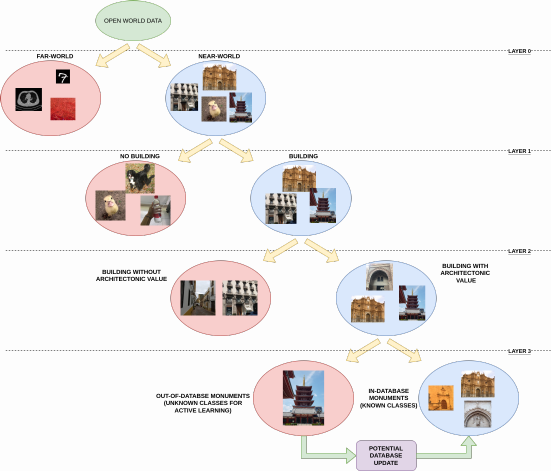}
    \caption{Schematic of the SeNeDiF-OOD architecture instantiated for the MonuMAI use case.}
    \label{fig:monumai_nested}
\end{figure}

Based on the previously defined worlds, the SeNeDiF-OOD architecture formulates a binary decision task between two consecutive categories. This section describes each fusion layer and analyzes the training strategy adopted at each level of the hierarchy. Figure \ref{fig:monumai_nested} illustrates, for the specific case of MonuMAI, the SeNeDiF-OOD instantiation and all its layers. Each of the layers will be outlined throughout this section.

\subsubsection{Layer 0 - Far-world filtering}

As layer zero of the SeNeDiF-OOD architecture, we propose an initial filtering stage designed to allow only images from the near-world to pass through, while discarding those that fall outside this domain. Since the far-world is inherently unbounded, whereas the near-world can be approximated using a dataset such as ImageNet, this layer employs an OOD detector specialized in far-OOD detection, that is, optimized for scenarios where the in-distribution (assumed here to be ImageNet) and the OOD distributions are markedly different.

Based on the study by Zhang et al. using the OpenOOD benchmark~\cite{zhang2023openood}, we selected the AdaScale-A postprocessor \cite{regmi2025adascaleadaptivescalingood} detector for this layer, due to its excellent performance in far-OOD scenarios when ImageNet serves as the in-distribution dataset. Moreover, the detector exhibits slightly reduced performance in near-OOD settings, meaning it may allow through samples visually similar to those in ImageNet when uncertainty arises. This behavior is not problematic at this stage, as it introduces a degree of tolerance that can later be refined by subsequent, more specialized filtering layers.

The detector was implemented using a RegNet-Y-16GF model pre-trained on ImageNet with Stochastic Weight Averaging-
Gaussian augmentation as the backbone architecture. Following our definition of the MonuMAI near-world as the domain of color images representing real-world scenarios, we configured the system to use a subset of ImageNet images as the in-distribution reference, since ImageNet comprises numerous categories that also include monuments and buildings, thereby aligning with the near-world domain. For MonuMAI far-world representation, we constructed a mixed out-of-distribution dataset comprising MNIST handwritten digits and DTD texture patterns, which represent inputs that do not meet the conditions of real-world color photography scenarios and thus fall outside the near-world domain.
The model extracts 3024-dimensional feature representations from the avgpool layer, and the AdaScale-A postprocessor applies adaptive scaling weights computed from class-wise variance statistics during the setup phase. The implementation employs standard ImageNet preprocessing with 384×384 pixel resolution and channel-wise normalization.  The optimal decision threshold is determined through F1-score maximization on a validation subset.

\subsubsection{Layer 1 - Building vs Non-Building}

Once the data have been filtered by layer zero, we can assume that the resulting domain corresponds to the near-world, consisting of images similar in nature to those in ImageNet. Within this constrained domain, the goal is to discriminate between buildings and non-buildings. In this case, the task can be effectively formulated as a binary classification problem.

For the non-building category, a subset of ImageNet images was selected, explicitly excluding categories related to buildings or architectural structures. For the building class, a diverse dataset was constructed by combining building-related images from ImageNet and, more extensively, outdoor architectural scenes from the \texttt{Places365} dataset~\cite{zhou2017places}, a large-scale scene recognition benchmark. A balanced selection was performed, resulting in a total of 50,000 images per class.

Based on this dataset, a small EfficientNet\_v2 model was trained to perform the binary classification between buildings and non-buildings.

\subsubsection{Layer 2 - Monument vs Non-Monument}

The second filtering layer is designed to distinguish monuments from buildings with no architectural relevance, assuming that only images of buildings are passed from the previous stage. This task is again framed as a binary classification problem, differentiating between architecturally significant monuments and generic buildings without historical or stylistic value. The main challenge in this layer lies in the construction of a dataset that captures sufficient diversity across both categories to ensure robust model generalization. To this end, a multi-stage pipeline was implemented to collect and curate the necessary data.

To assemble this corpus, we relied on Wikimedia Commons as the primary source and implemented a single end-to-end pipeline that begins with large-scale category exploration and culminates in a CLIP-based semantic screening. We first traversed Commons categories aligned with cultural heritage and generic architecture, recursively exploring their subcategory trees to harvest image URLs and metadata (e.g., page titles and category paths). Because category annotations are noisy and often include off-topic material, such as interiors, monument pages, maps, drawings, plans, or close-up artifacts, we treated the raw download as a broad superset. A light preprocessing stage standardized file formats and sizes, enforced minimum resolution and aspect-ratio constraints, and removed near-duplicates using perceptual hashes, yielding a cleaner pool for semantic verification.

To ensure that retained images match their intended concept, we applied CLIP-based thematic filtering with a multi-prompt strategy. For each target concept, we authored several complementary text prompts and embedded both images and prompts with CLIP. We computed cosine similarities between each image embedding and all prompt embeddings and took the maximum over prompts as a robust match score. An image was accepted if this maximum score was at least $\tau=0.25$, a threshold tuned on a small, human-verified validation split; otherwise, it was relegated to an unmatched pool. This multi-prompt procedure mitigates the brittleness of single prompts and reduces false rejections, enabling scalable, largely automated thematic curation.

After filtering, we performed a static train/validation/test split that preserves geographic and stylistic diversity, and we mapped the harvested categories to the layer-specific binary task: \emph{monuments} (aggregating both known and unknown styles) versus \emph{non-monument buildings}. Using the curated data, we trained a ResNet-50 initialized with ImageNet pretrained weights in a transfer-learning setup for binary classification between monuments and non-monument.

\subsubsection{Layer 3 - Known vs Unknown Style}

Once the domain has been narrowed to monument images, the final filtering stage aims to detect those images that correspond to the architectural styles recognized by MonuMAI.

This stage is formulated as an OOD detection problem over the MonuMAI model itself, which has already been trained on its corresponding in-distribution dataset. Additionally, the monument dataset constructed for the previous layer can be leveraged to support outlier exposure, thereby facilitating the discrimination between known and unknown architectural styles.

As in previous layers, we rely on the OpenOOD benchmark to select a suitable detector. In this case, the setting corresponds to a near-OOD scenario, and the dataset is entirely custom-built, unlike in layer zero, where ImageNet was used as the in-distribution class. For this reason, we selected entropy-based detection with outlier exposure fine-tuning ~\cite{hendrycks2019deepanomalydetectionoutlier}~\cite{hendrycks2018baselinedetectingmisclassifiedoutofdistribution}, a detector that demonstrates strong performance in near-OOD detection across multiple in-distribution datasets and supports training with outlier exposure.

The implementation leverages the pre-trained MonuMAI model, which was originally trained to classify four distinct Iberian architectural styles. Rather than using the model in its frozen state, we perform fine-tuning with a composite loss function that incorporates outlier exposure principles. The training procedure combines two complementary objectives: preserving the model's existing knowledge of known architectural styles while explicitly learning to produce uniform probability distributions for unknown styles.

Specifically, the outlier exposure training employs a mixed loss function:

\begin{equation*}
\mathcal{L}_{\text{total}} = \mathcal{L}_{\text{in-distribution}} + \lambda \times \mathcal{L}_{\text{OE}}
\end{equation*}

where $\mathcal{L}_{\text{in-distribution}}$ represents the standard cross-entropy loss computed on in-distribution samples (known architectural styles from the MonuTrain dataset), and $\mathcal{L}_{\text{OE}}$ implements the outlier exposure loss that encourages uniform probability distributions over unknown architectural styles. The outlier exposure loss is formulated as a KL divergence between the model's predicted probabilities for out-of-distribution samples and a uniform distribution:

\begin{equation*}
\mathcal{L}_{\text{OE}} = \text{KL}\left(\text{softmax}(f(x_{\text{outlier}})) \parallel \text{Uniform}\left(\frac{1}{K}\right)\right)
\end{equation*}

where $K=4$ represents the number of architectural classes in the original MonuMAI model. The hyperparameter $\lambda=0.5$ balances the preservation of existing knowledge with the acquisition of OOD detection capabilities.

For training, we utilize a custom dataset of monument images constructed through the Wikimedia Commons scraping procedures and CLIP-based thematic filtering described in previous sections (Layer 2), comprising architectural styles from across the globe that are not present in the MonuMAI in-distribution training dataset. These globally diverse monument images serve as outlier samples, ensuring that the model learns to recognize architectural styles beyond those found in the Iberian Peninsula.

At inference time, OOD detection is performed using entropy-based scoring. Images corresponding to known Iberian architectural styles produce low-entropy predictions (high confidence in one of the four known classes), while images depicting unknown architectural styles generate high-entropy predictions (uniform uncertainty across all classes). The detection threshold is calibrated using the 95th percentile of entropy scores computed on known style samples, ensuring robust discrimination between in-distribution and out-of-distribution architectural content.

\section{Experimental analysis} \label{sec:experiments}

This section presents the experimental analysis conducted to evaluate the SeNeDiF-OOD methodology using the MonuMAI model as a case study. First, we describe the datasets employed and the procedure followed during experimentation. Subsequently, we report and analyze the results.

\subsection{Experimental setup}

The first step in providing an appropriate performance measure for OOD detection in the context of architectural style classification is the construction of a sufficiently broad and diverse test dataset. This dataset must include a representative number of images from all architectural styles recognized by MonuMAI, as well as from all OOD categories defined within the SeNeDiF-OOD methodology.

To this end, we created the MonuTest-OOD dataset, which was manually collected and curated from multiple sources. The primary source consisted of images uploaded directly to the MonuMAI server. These contributions included both authentic photographs of monuments in different styles, encompassing those recognized and unrecognized by the model, and OOD images uploaded by users. The latter group comprised sculptures and artworks not classified as monuments, buildings without architectural relevance, and miscellaneous images unrelated to the application domain. To ensure sufficient variability across all categories, the dataset was complemented with additional images retrieved from Wikimedia.

In total, MonuTest-OOD contains 6354 images distributed across the different OOD and in-distribution categories. The detailed category distribution is provided in Table~\ref{tbl:monutest}.

\begin{table}[H]
\centering
\caption{Distribution of images in the MonuTest-OOD dataset. The column ``Number of exclusive images'' shows the number of images in that category not belonging to any subcategory.}
\label{tbl:monutest}
\renewcommand{\arraystretch}{1.2} 
\begin{tabular}{l c c}
\hline
\textbf{Category} & \textbf{Number of exclusive images}\\
\hline
Far-world (OOD)    & 1550 \\
Near-world (OOD)   & 1518 \\
Buildings (OOD)    & 1467 \\
Unknown Monuments (OOD)    & 781 \\
Known monuments (in-distribution) & 1038 \\
\hline
\textbf{Total}  \textbf{(in-distribution + OOD)} & \textbf{6354} \textbf{(1038 + 5316)}\\
\hline
\end{tabular}
\end{table}

The main objective of the experimentation is to analyze the performance of the SeNeDiF-OOD methodology, trained as described in Section~\ref{sec:nested_monumai} and using MonuTest-OOD as the evaluation dataset. To this end, the layers of the system will be analyzed both individually and globally, and metrics such as AUROC, F1 score, overall accuracy, or per-class accuracy rates will be reported.

The results of the SeNeDiF-OOD methodology will be compared with those obtained by MonuMAI itself in OOD detection. Although MonuMAI is not explicitly designed for this purpose, it incorporates a mechanism that allows rejecting images that are not related to monuments. The MonuMAI-KED model, which is part of the pipeline and responsible for identifying architectural elements in the image, is an object detection model. Consequently, there are cases in which no elements are detected in the image. In such situations, the MonuMAI pipeline filters the input as OOD, regardless of the architectural style that might have been assigned by the MonuNet network. The main limitation of this OOD filtering strategy lies in the tendency of MonuMAI-KED to interpret certain patterns as architectural elements even when the image itself is unrelated to monuments. As a result, the effectiveness of this filtering approach is severely limited.

Additionally, a qualitative analysis will also be conducted on the images filtered at each layer of the system, with the aim of drawing relevant conclusions regarding the individual behavior of each layer. The OOD images misclassified as monuments will also be analyzed and compared between the SeNeDiF-OOD methodology and the MonuMAI original OOD pipeline.

\FloatBarrier

\subsection{Metrics and evaluation protocol}

We adopt five standard metrics to quantify performance at each gate and for the end-to-end system. Let $\mathrm{TP}$, $\mathrm{FP}$, $\mathrm{TN}$ and $\mathrm{FN}$ denote true positives, false positives, true negatives and false negatives, respectively, with the positive class defined as indicated below.

\begin{itemize}
  \item \textbf{Accuracy} $\displaystyle = \frac{\mathrm{TP}+\mathrm{TN}}{\mathrm{TP}+\mathrm{FP}+\mathrm{TN}+\mathrm{FN}}$ \\ \\ Overall fraction of correct decisions. It is informative but can be influenced by class imbalance.

  \item \textbf{Precision} $\displaystyle = \frac{\mathrm{TP}}{\mathrm{TP}+\mathrm{FP}}$ \\ \\ Among the samples predicted as positive, the fraction that are truly positive. In our deployment, it measures how clean are the images the system accepts as in-distribution.

  \item \textbf{Recall} $\displaystyle = \frac{\mathrm{TP}}{\mathrm{TP}+\mathrm{FN}}$ \\ \\ Among truly positive samples, the fraction correctly identified as positive. It captures how many relevant images the system retains rather than rejects.

  \item \textbf{F1-score} $\displaystyle = \frac{2\cdot \mathrm{Precision}\cdot \mathrm{Recall}}{\mathrm{Precision}+\mathrm{Recall}}$ \\ \\ Harmonic mean that balances precision and recall, useful when their trade-off matters. \\ \\

  \item \textbf{AUC (AUROC)}. Area under the Receiver Operating Characteristic curve obtained by sweeping a decision threshold over a real-valued score. It is threshold-free and reflects separability between positive and negative classes.
\end{itemize}

Each gate is a binary problem with its own positive class:
\begin{itemize}
  \item \textbf{Layer 0 (far-world filter):} positive = near-world pass candidates; negative = far-world.
  \item \textbf{Layer 1 (building check):} positive = buildings; negative = non-buildings.
  \item \textbf{Layer 2 (monument check):} positive = monuments (known + unknown); negative = buildings.
  \item \textbf{Layer 3 (style knownness):} positive = known-style monuments; negative = unknown-style monuments.
\end{itemize}

Per-layer Accuracy, Precision, Recall and F1 are computed at the operating threshold chosen for that gate (calibrated on a small held-out split). Per-layer AUC is computed by sweeping that gate’s score (e.g., confidence/energy or feature-distance) on the corresponding test subset.

For the end-to-end evaluation, we treat the cascade as a single classifier with \emph{known-style monument} as the positive class and \emph{all other categories} as negative. An image counts as a positive prediction only if it passes layers~0–2 and layer~3 predicts a known style. Accuracy, Precision, Recall, and F1 are computed directly from these final decisions. For AUC, we use the calibrated acceptance score at layer~3; samples rejected upstream receive the minimum acceptance score so that the ROC reflects the whole cascade.

In the context of our work, Precision quantifies the risk of admitting non-target content, while Recall quantifies how many true monuments of known style are preserved. F1 summarizes their balance. AUC indicates how robustly a gate (or the final system) separates positives from negatives regardless of the exact threshold. In cultural-heritage use, conservative operation typically prioritizes high Precision, accepting a moderate drop in Recall; we report both to make this trade-off explicit.


\subsection{Results}

This section presents the results obtained from the different experiments conducted.

\subsubsection{Results by layers}

Tables \ref{tbl:layer0}, \ref{tbl:layer1}, \ref{tbl:layer2_accuracy}, \ref{tbl:layer2} and \ref{tbl:layer3} present the results of layers 0, 1, 2, and 3, respectively, of the SeNeDiF-OOD methodology applied to MonuMAI. The metrics reported correspond to the binary problem defined at each layer.

\begin{table}[!htbp]
\centering
\caption{Results for layer 0 (far-world filtering) of the SeNeDiF-OOD methodology.}
\label{tbl:layer0}
\begin{tabular}{@{}ccccccc@{}}
\toprule
AUROC & Accuracy & Precision & Recall (in-distribution) & F1 & Specificity OOD & FPR95 \\
\midrule
0.9601 & 0.9397 & 0.9076 & 0.9776 & 0.9413 & 0.9026 & 0.0974 \\
\bottomrule
\end{tabular}
{
\captionsetup{justification=centering,singlelinecheck=false}
\caption*{\footnotesize \textbf{In-Distribution class:} near-world images (pass candidates).\\
\textbf{OOD class:} far-world images (rejected).\\
\textit{Note:} FPR95 computed at 95\% TPR on in-distribution class. Precision and F1 are derived from (Accuracy, Recall, Specificity) under the test-set class prior.}
}
\end{table}

\begin{table}[!htbp]
\centering
\caption{Results for layer~1 (building vs no-building) of the SeNeDiF-OOD methodology.}
\label{tbl:layer1}
\begin{tabular}{cccccc}
\toprule
ACC & PREC & F1 & Recall & AUROC \\
\midrule
0.9266 & 0.8878 & 0.9314 & 0.9796 & 0.9774 \\
\bottomrule
\end{tabular}
{
\captionsetup{justification=centering,singlelinecheck=false}
\caption*{\footnotesize \textbf{In-Distribution class:} buildings.\\
\textbf{OOD class:} non-buildings.}
}
\end{table}

\begin{table}[!htbp]
\centering
\caption{Overall performance for layer~2 (monument vs. building classification) of the SeNeDiF-OOD methodology by subcategory.}
\label{tbl:layer2_accuracy}
\begin{tabular}{lccc}
\toprule
Subcategory & Samples & Accuracy & AUROC \\
\midrule
All                & 3286 & 0.9394 & 0.9502 \\
\midrule
Known Monuments    & 1038 & 0.9422 & \textemdash \\
Unknown Monuments  & 781  & 0.9181 & \textemdash \\
Buildings          & 1467 & 0.9489 & \textemdash \\
\bottomrule
\end{tabular}
{
\captionsetup{justification=centering,singlelinecheck=false}
\caption*{\footnotesize \textbf{In-Distribution class:} monuments (known + unknown).\\
\textbf{OOD class:} buildings.\\
\textit{Note:} AUROC is computed globally for the binary decision (monument vs.\ building); per-subcategory AUROC is not applicable.}
}
\end{table}

\begin{table}[!htbp]
\centering
\caption{Detailed results for layer~2 (monument vs building classification) of the SeNeDiF-OOD methodology.}
\label{tbl:layer2}
\begin{tabular}{llcccc}
\toprule
Scope & Subcategory & Class & Precision & Recall & F1 \\
\midrule
\multirow{2}{*}{Global} & \multirow{2}{*}{All (3286 samples)} & Monument & 0.9576 & 0.9318 & 0.9446 \\
                        &                                      & Building & 0.9182 & 0.9489 & 0.9333 \\
\midrule
\multirow{1}{*}{Per-subcategory} & \multirow{1}{*}{Known Monuments (1038)}    & Monument & 1.0000 & 0.9422 & 0.9702 \\
\multirow{1}{*}{Per-subcategory} & \multirow{1}{*}{Unknown Monuments (781)}  & Monument & 1.0000 & 0.9181 & 0.9573 \\
\multirow{1}{*}{Per-subcategory} & \multirow{1}{*}{Buildings (1467)}         & Building & 1.0000 & 0.9489 & 0.9738 \\
\bottomrule
\end{tabular}
{
\captionsetup{justification=centering,singlelinecheck=false}
\caption*{\footnotesize \textbf{In-Distribution class:} monuments (known + unknown).\\
\textbf{OOD class:} buildings.\\
\textit{Note:} In class-wise rows, metrics treat the shown class as the positive label.}
}
\end{table}

\begin{table}[!htbp]
\centering
\caption{Results for layer~3 (known vs unknown monuments) of the SeNeDiF-OOD methodology.}
\label{tbl:layer3}
\begin{tabular}{ccccccc}
\toprule
ACC & PREC & F1 & Recall & AUROC & AUPR \\
\midrule
0.9245 & 0.7851 & 0.7983 & 0.8120 & 0.9660 & 0.8353 \\
\bottomrule
\end{tabular}
{
\captionsetup{justification=centering,singlelinecheck=false}
\caption*{\footnotesize \textbf{In-Distribution class:} known-style monuments.\\
\textbf{OOD class:} unknown-style monuments.}
}
\end{table}

\begin{figure}[htbp]
    \centering
    \includegraphics[width=\textwidth]{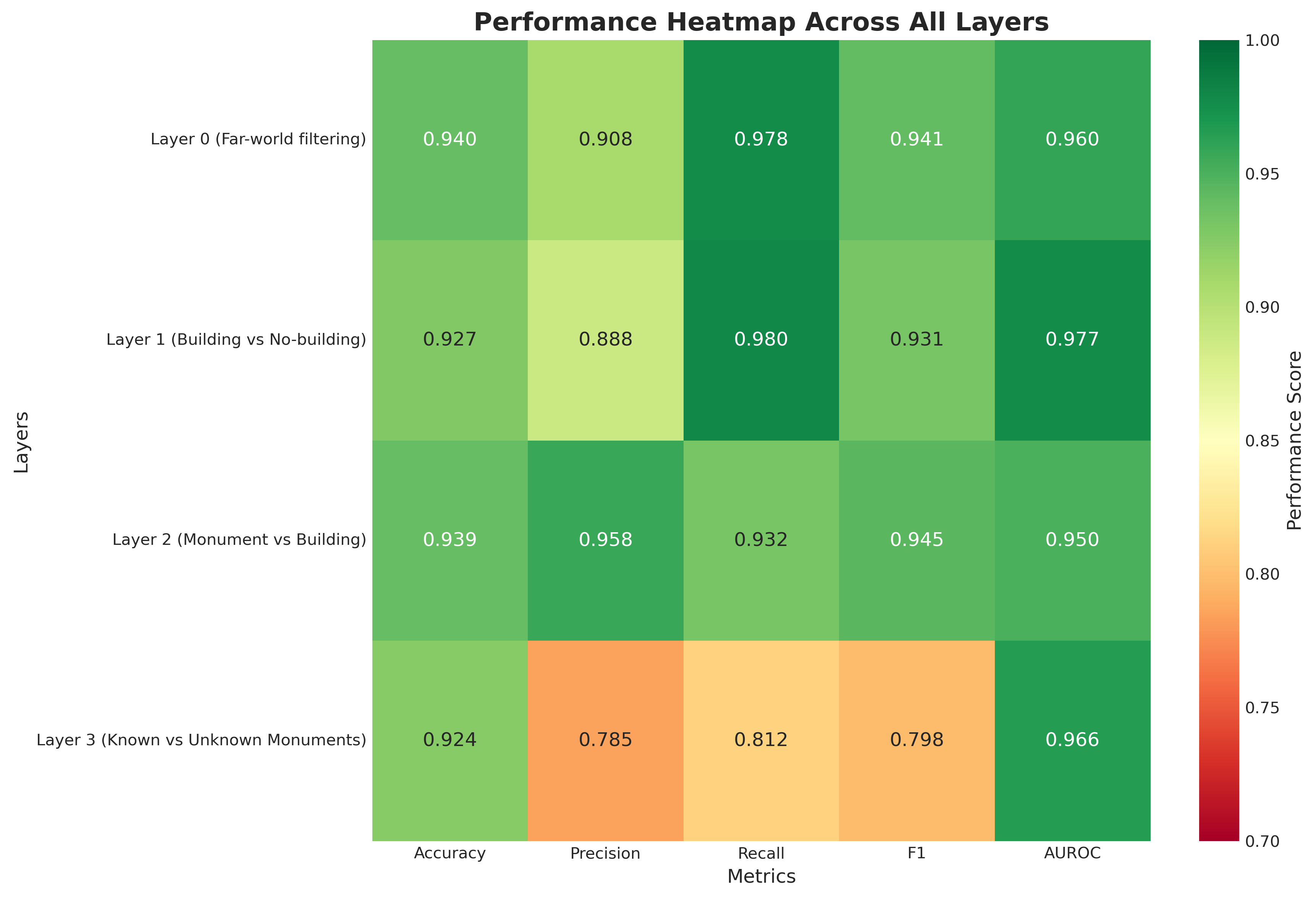}
    \caption[Layer-wise performance heatmap]{Layer-wise performance heatmap 
    across the main metrics (Accuracy, Precision, Recall, F1, AUROC). 
    Each row corresponds to a gate in the SeNeDiF-OOD pipeline 
    (far-world filter; building vs.\ no-building; monument vs.\ building; 
    known vs.\ unknown style).}
    \label{fig:layer-heatmap}
\end{figure}
\begin{figure}[htbp] 
    \centering
    
    \begin{subfigure}[b]{0.49\textwidth}
        \centering
        \includegraphics[width=\textwidth]{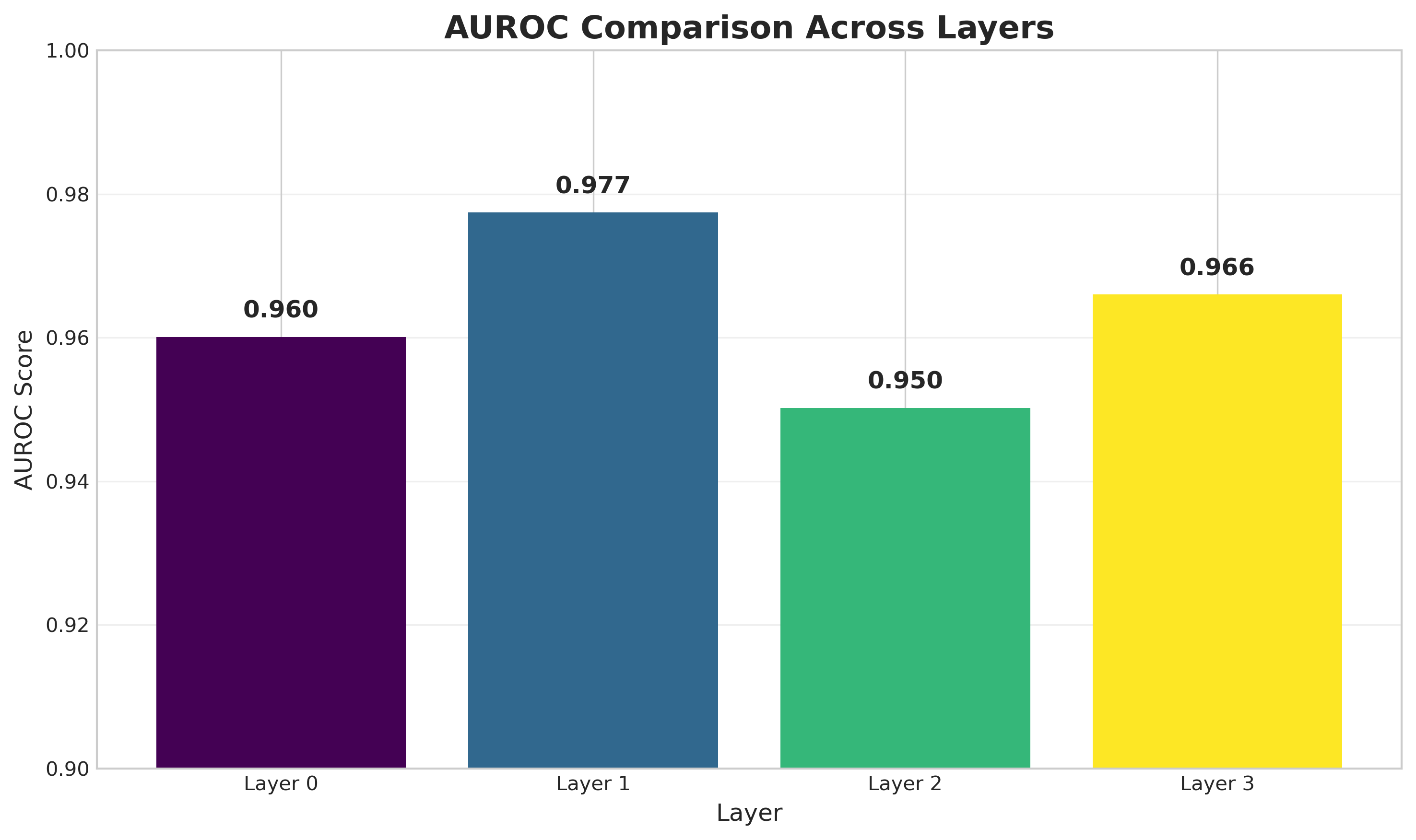}
        \caption{AUROC comparison across layers.}
        \label{fig:layer-auroc} 
    \end{subfigure}
    \hfill 
    \begin{subfigure}[b]{0.49\textwidth}
        \centering
        \includegraphics[width=\textwidth]{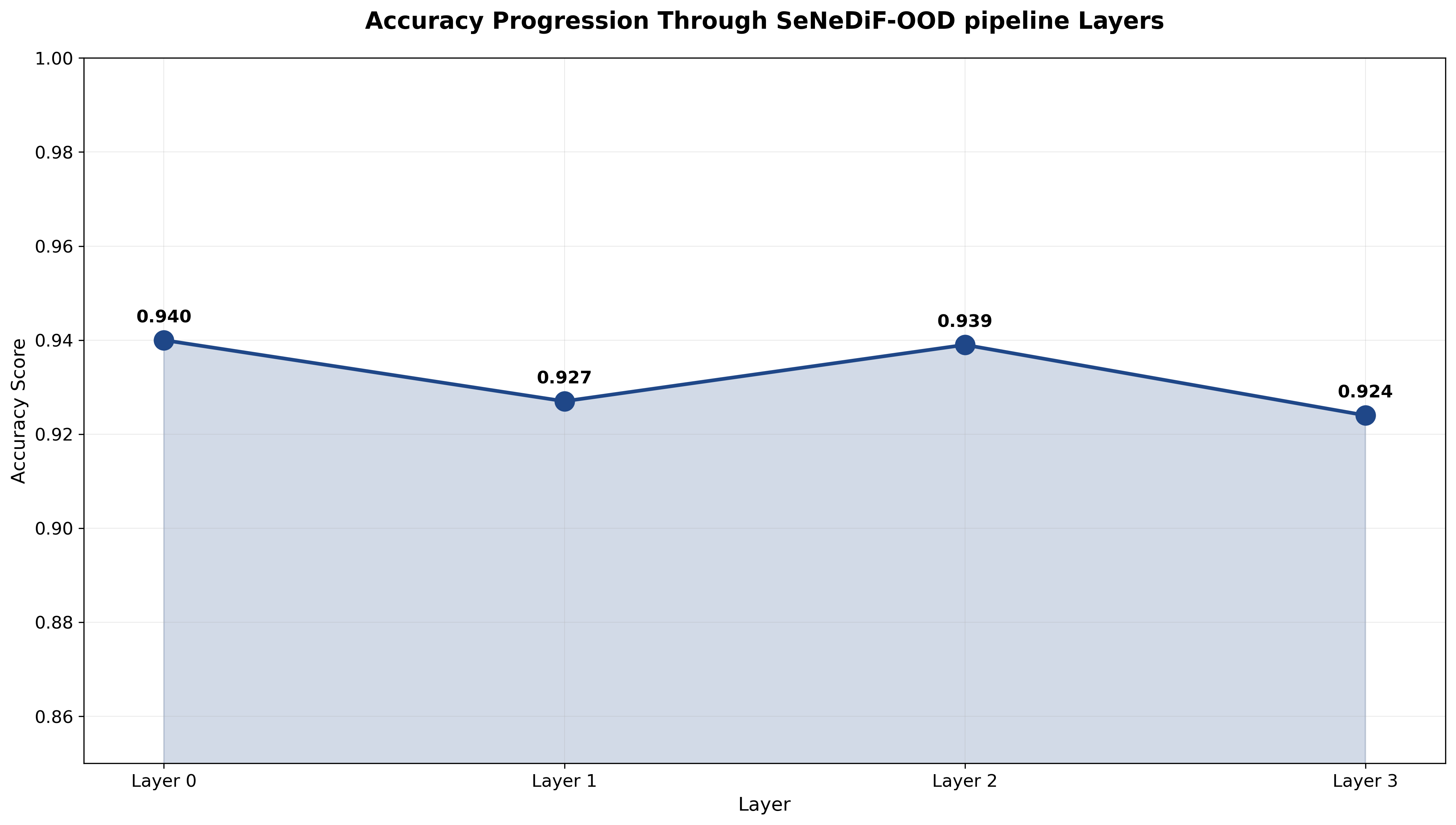}
        \caption{Accuracy progression through the hierarchy.}
        \label{fig:layer-accuracy} 
    \end{subfigure}
    
    \caption{Layer-wise performance trends. (a) AUROC comparison, 
    emphasizing separability at each gate. (b) Accuracy progression 
    as decisions approach fine-grained boundaries.}
    \label{fig:layer-trends} 
\end{figure}

To complement the tables, we include visual summaries that condense 
the layer-wise behavior of the pipeline. \autoref{fig:layer-heatmap} 
offers an at-a-glance view of the main metrics across all gates. 
Focusing on separability and accuracy trends, \autoref{fig:layer-trends} 
compares the AUROC per layer and traces the accuracy 
progression along the hierarchy.

\begin{figure}[!htbp]
  \centering
  \includegraphics[width=\linewidth]{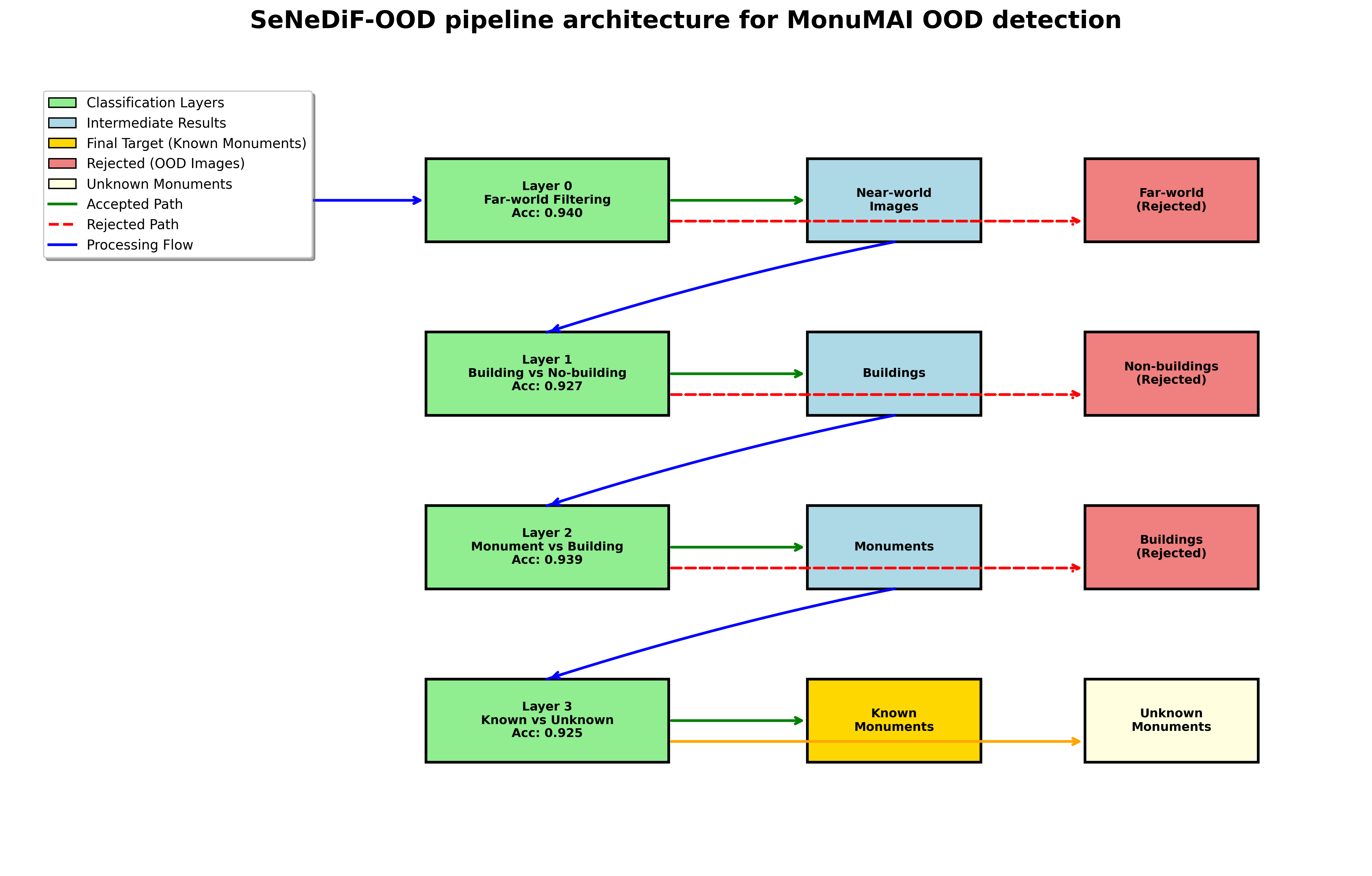}
  \caption[Architecture overview of the \textcolor{orange}{SeNeDiF-OOD pipeline}]{Architecture overview of the SeNeDiF-OOD pipeline used for OOD detection on MonuMAI.} 
  \label{fig:arch-overview}
\end{figure}

Figure \autoref{fig:arch-overview} summarizes the end-to-end flow of the SeNeDiF-OOD pipeline. The diagram shows the four binary gates, the accepted and rejected paths at each decision, and the measured accuracy per layer on the test split. This view links the metrics to the operational behavior of the system.


\subsubsection{Global results}

We report end to end OOD performance on \textit{MonuTest-OOD} using \emph{known-style monument} as the positive class for all metrics. In the cascade, the candidate pool for the positive class narrows progressively through the gates: an image must pass the far-world filter (layer~0), be accepted as a building (layer~1), be accepted as a monument (layer~2), and finally be assigned to one of the known styles (layer~3). An image is counted as a positive prediction only if it survives layers~0--2 and layer~3 predicts a known style; otherwise it is counted as negative. Table~\ref{tbl:global_results} summarizes the resulting end-to-end metrics. The results are compared with those of the original MonuMAI OOD system, which classifies an input as OOD when no architectural elements are detected in the image.

\begin{table}[!htbp]
\centering
\caption{Global results for OOD detection on the MonuTest-OOD dataset (positive class: known-style monument).}
\label{tbl:global_results}
\renewcommand{\arraystretch}{1.2}
\begin{tabular}{lccccc}
\toprule
\textbf{Model} & \textbf{Accuracy} & \textbf{Precision} & \textbf{Recall} & \textbf{F1-score} & \textbf{AUROC} \\
\midrule
SeNeDiF-OOD & \textbf{0.9417} & \textbf{0.8823} & 0.7625 & \textbf{0.8180} & \textbf{0.9642} \\
MonuMAI Pipeline   & 0.7748 & 0.4130 & \textbf{0.8991} & 0.5660 & 0.9268 \\
\bottomrule
\end{tabular}
{
\captionsetup{justification=centering,singlelinecheck=false}
\caption*{\footnotesize \textbf{in-distribution class:} known-style monuments.\\
\textbf{OOD class:} all other categories (unknown-style monuments, buildings, non-buildings, far-world).}
}
\end{table}

\FloatBarrier

Each test image was routed through the four-gate hierarchy. We recorded the final \emph{accept} or \emph{reject} decision and a calibrated acceptance score at the last gate. Metrics in Table~\ref{tbl:global_results} are computed on these final outputs with \emph{known-style monument} as the positive class and \emph{all other} inputs as the negative class, which mirrors deployment: an image counts as positive only if it passes gates~0--2 and layer~3 predicts a known style. For AUC, we sweep a threshold over the last-gate acceptance score, where samples rejected upstream are assigned the minimum score so that the ROC reflects the whole cascade.

The cascade attains high accuracy (0.9417) and strong separability (AUC 0.9642). Precision is also high (0.8823), meaning that accepted monuments are rarely false positives. Recall is lower (0.7625), which is expected in a conservative design: upstream thresholds and the known versus unknown style boundary prefer to reject ambiguous cases rather than admit non–monuments or unfamiliar styles. With an OOD–heavy test set, this choice naturally lifts accuracy while reducing recall. In general, the ideal operating point is a balance between precision and recall that depends on the application's risk tolerance and user goals. In our cultural-heritage setting and with the presented configuration, we effectively suppress the clear OOD cases that harmed the original MonuMAI pipeline, where totally unrelated non-monument images slipped through as monuments (see \autoref{fig:OOD_examples} for reference). This mitigation is the primary practical gain of the cascade; the recall deficit largely concentrates on borderline positives (partial views, occlusions, non-canonical shots), which can be recovered with targeted calibration or data regularization.

\begin{figure}[!htbp]
  \centering
  \includegraphics[width=\linewidth]{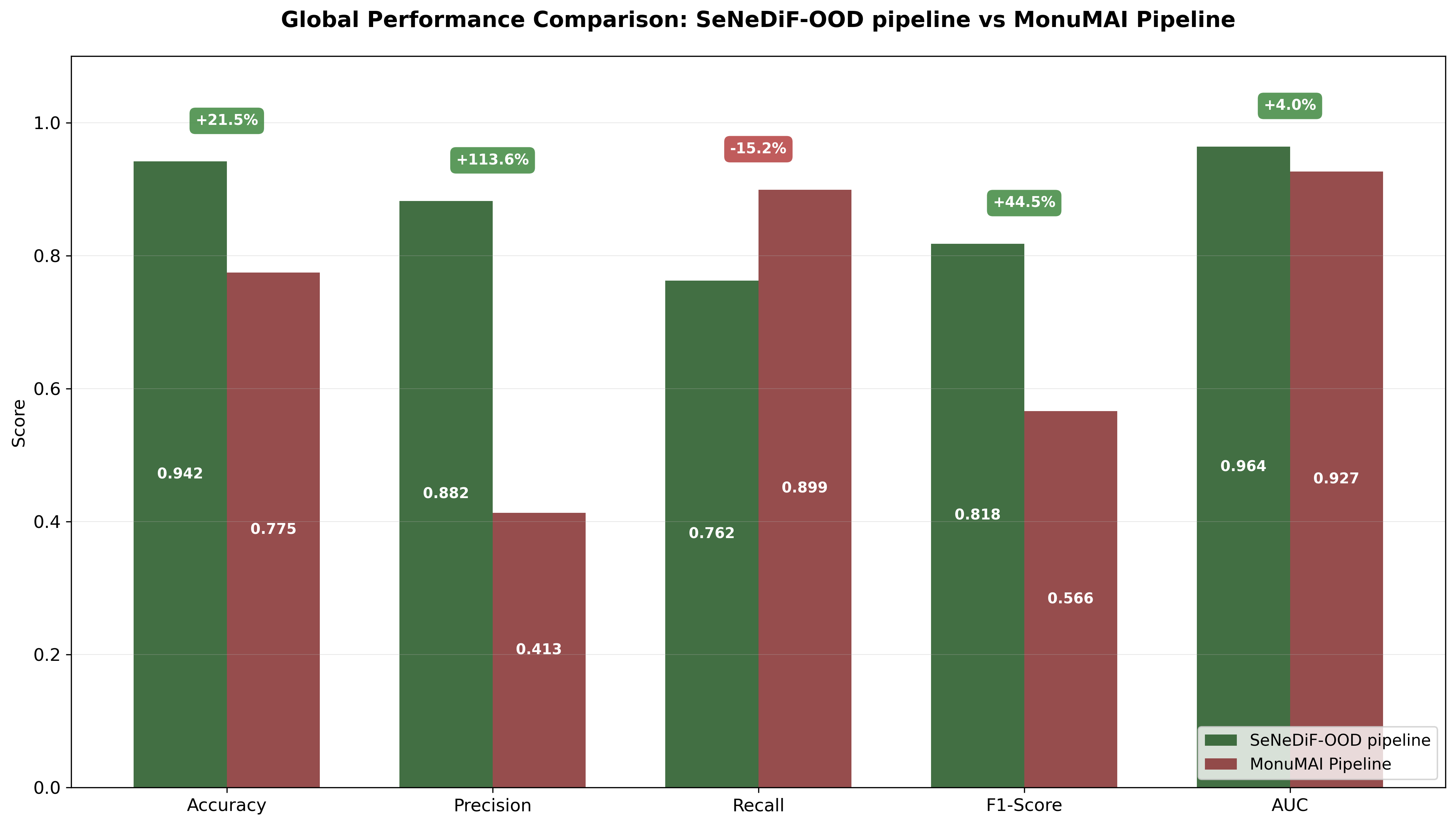}
  \caption[Global performance comparison]{Global performance comparison on \emph{MonuTest-OOD} (positive class: progressive semantic narrowing of known-style monuments class across the cascade). Bars show Accuracy, Precision, Recall, F1, and AUC for the SeNeDiF-OOD pipeline versus the original MonuMAI pipeline. Annotations emphasize relative changes in each metric, underscoring the reduction of false positives without sacrificing significant coverage of in-distribution monuments.}
  \label{fig:global-visual}
\end{figure}

Beyond tabular reporting, Figure~\ref{fig:global-visual} contrasts end–to–end metrics between our SeNeDiF-OOD pipeline and the original MonuMAI filtering. The visualization highlights the precision gains that reduce false positives while maintaining useful recall and excellent AUC, which aligns with the goal of preserving known–style monuments and rejecting OOD inputs.

\paragraph{On the precision-recall tradeoff and the scope of application.}

While the present configuration is intentionally precision-leaning for cultural-heritage use, the same cascade can be tuned to different risk profiles in other application areas. The precision–recall balance is not universal; it depends on downstream costs, user goals, and governance constraints. In the following, we discuss how this concept may apply to several other domains.

\begin{itemize}

  \item \textbf{Clinical and hospital workflows.} The preferred operating point varies by task. In screening and triage, many teams emphasize high recall to avoid missing disease, accepting a moderate number of false alarms if follow-up is cheap and safe. In decision-support that may trigger invasive or high-risk interventions, institutions often minimize false positives to reduce harm and alarm fatigue. A layered system like ours can support either policy: upstream gates can be tuned for high recall, with stricter confirmatory gates downstream, or the reverse when specificity is paramount~\cite{hendrycks2021unsolved,zamzmi2024out}.

  \item \textbf{Autonomous driving and safety monitoring.} For hazard detection, failing to alert can be catastrophic, so recall is usually prioritized at early stages, while later gates verify to keep false positives within acceptable rates~\cite{henriksson2023evaluation}. Layer-specific thresholds and validation splits make these choices explicit and auditable.

  \item \textbf{Fraud and abuse detection.} Blocking legitimate activity can be costly and erode trust, so precision is often favored. Class-conditional thresholds and cost-sensitive calibration can bias the final gate toward very low false positives while keeping recall at acceptable levels.
\end{itemize}

\noindent The SeNeDiF-OOD architecture makes these adaptations straightforward. Per-gate thresholds can be calibrated on held-out data to meet domain targets; class-conditional thresholds can correct systematic conservatism for specific classes; and the strength of outlier exposure can be adjusted to widen or narrow acceptance regions near subtle boundaries. Reporting precision–recall curves together with AUC, and attributing where true positives are lost along the cascade, helps select an operating point that matches the risk profile of each domain~\cite{zhang2023openood,yang2024generalized}.


\subsection{Images filtered by layers}

This section provides a qualitative, image-based view of how the cascade behaves in practice. Here, we examine concrete inputs that either slip through a layer or that progressively survive all filters. The goal is twofold: first, to distill characteristic visual patterns that explain the few errors that remain after each gate, and second, to illustrate how the hierarchical fusion cascade progressively narrows the operational design domain from generic imagery to known-style monuments. These observations will be revisited in the discussion and used to motivate targeted data curation and training refinements.

\subsubsection{Representative misclassifications per layer}

\begin{figure}[!htbp]
\centering
\footnotesize

\textbf{Layer 0: near/far-OOD textures that slipped through the far-world filter.}\\[0.25em]

\begin{subfigure}[t]{0.3\textwidth}
  \centering
  \includegraphics[height=0.13\textheight]{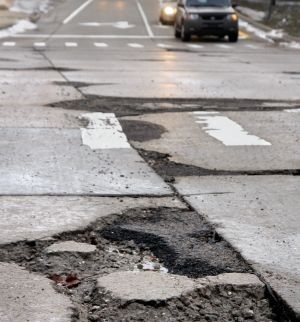}
  \caption{Texture: \emph{potholed}}
\end{subfigure}\hfill
\begin{subfigure}[t]{0.3\textwidth}
  \centering
  \includegraphics[height=0.13\textheight]{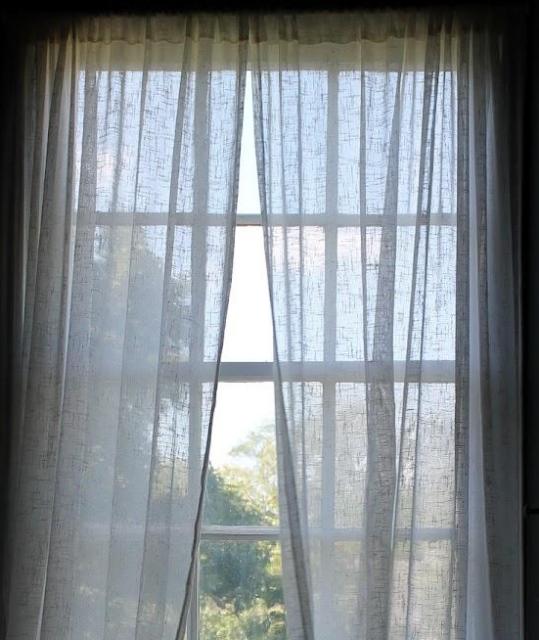}
  \caption{Texture: \emph{gauzy}}
\end{subfigure}\hfill
\begin{subfigure}[t]{0.3\textwidth}
  \centering
  \includegraphics[height=0.13\textheight]{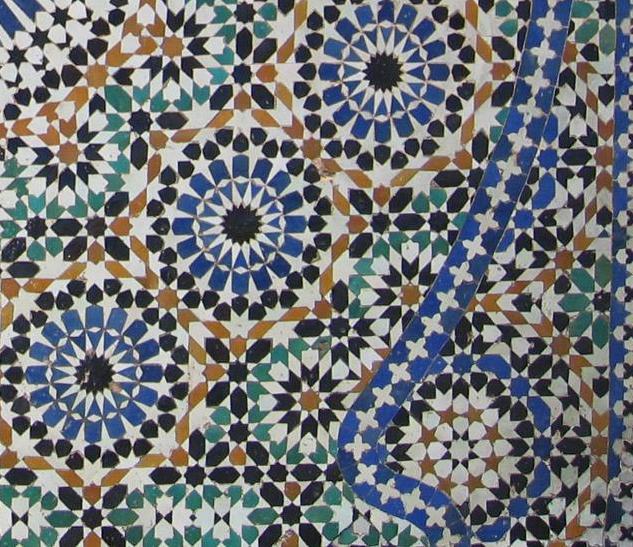}
  \caption{Texture: \emph{interlaced}}
\end{subfigure}

\par\vspace{0.7em}

\textbf{Layer 1: building/non-building errors despite the near-world gate.}\\[0.25em]

\begin{subfigure}[t]{0.3\textwidth}
  \centering
  \includegraphics[height=0.13\textheight]{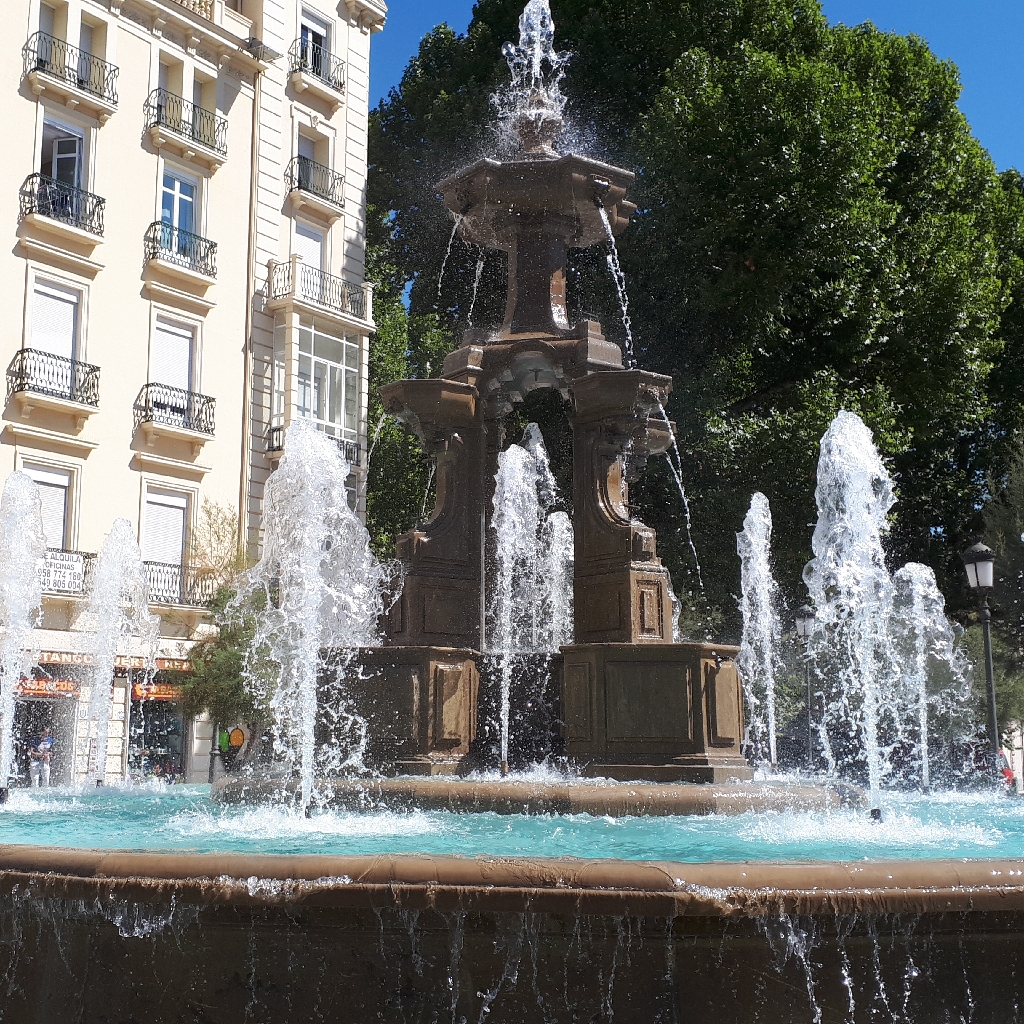}
  \caption{Fountain accepted as building}
\end{subfigure}\hfill
\begin{subfigure}[t]{0.3\textwidth}
  \centering
  \includegraphics[height=0.13\textheight]{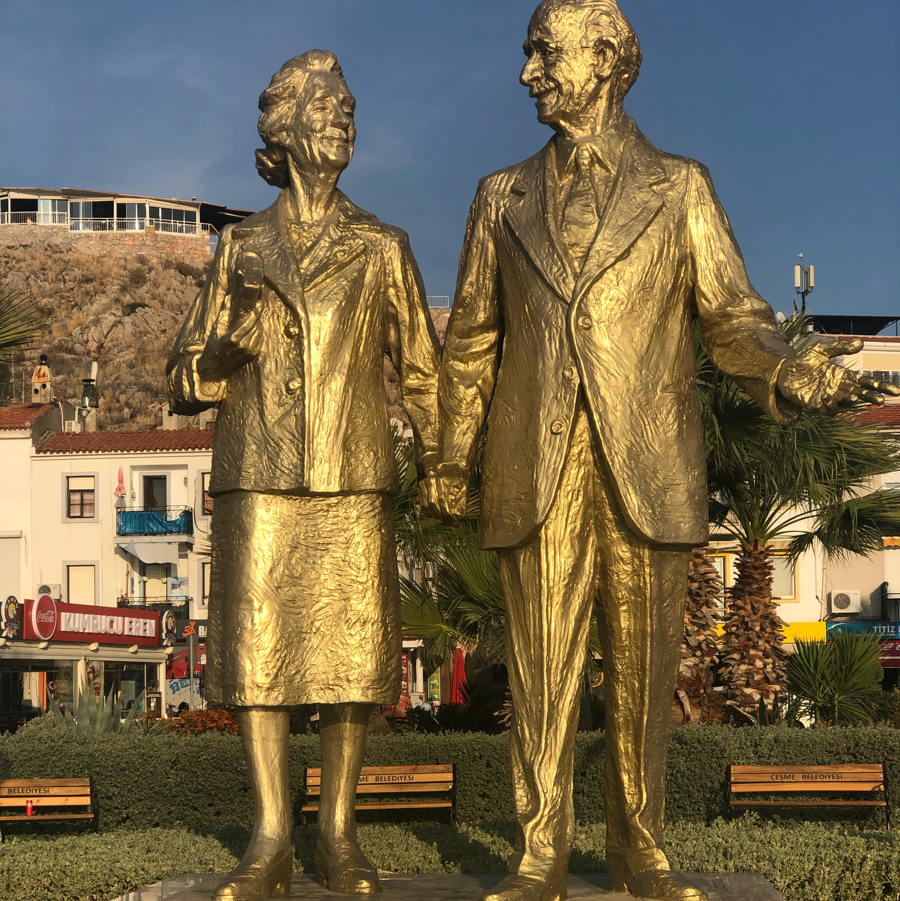}
  \caption{Statues accepted as building}
\end{subfigure}\hfill
\begin{subfigure}[t]{0.3\textwidth}
  \centering
  \includegraphics[height=0.13\textheight]{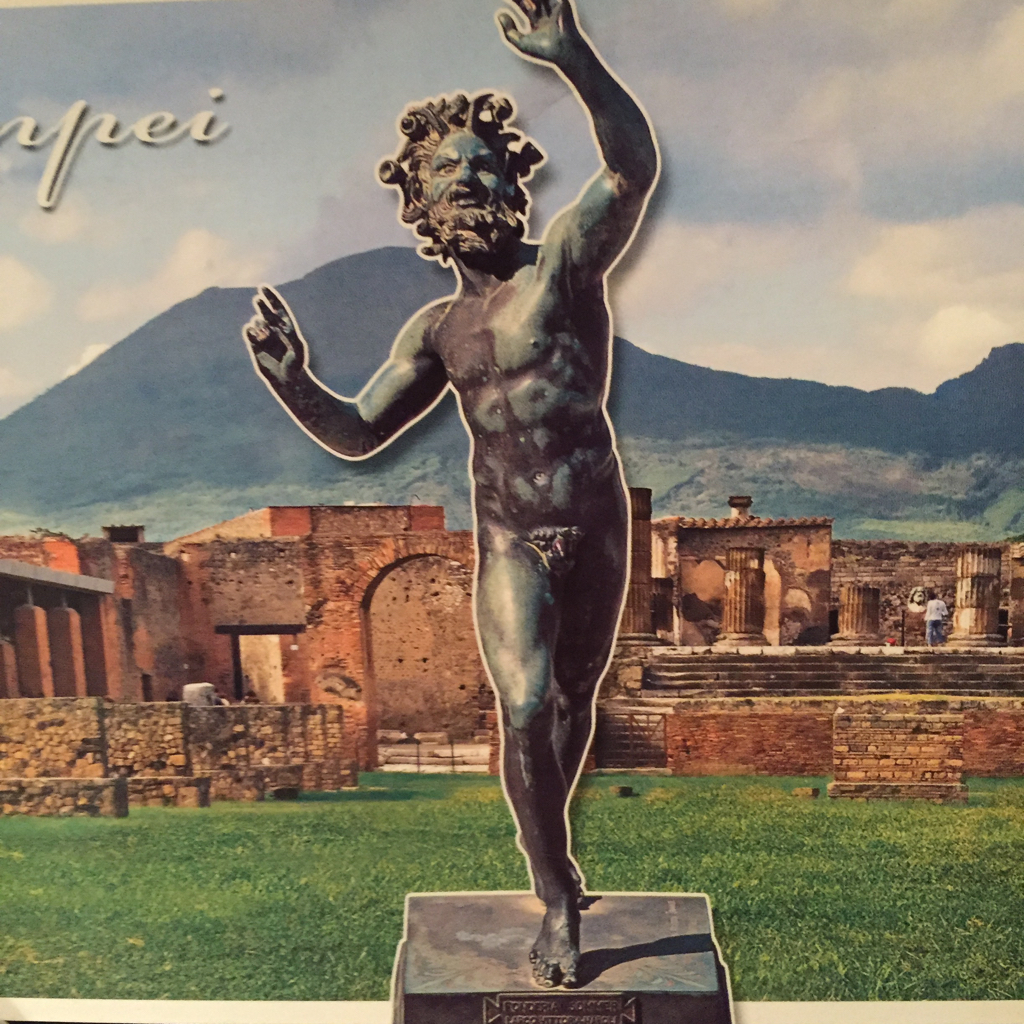}
  \caption{Edge case of a drawing of a figure accepted as building}
\end{subfigure}

\par\vspace{0.7em}

\textbf{Layer 2: monument vs. non-monument errors where generic buildings mimic monument cues.}\\[0.25em]

\begin{subfigure}[t]{0.3\textwidth}
  \centering
  \includegraphics[height=0.13\textheight]{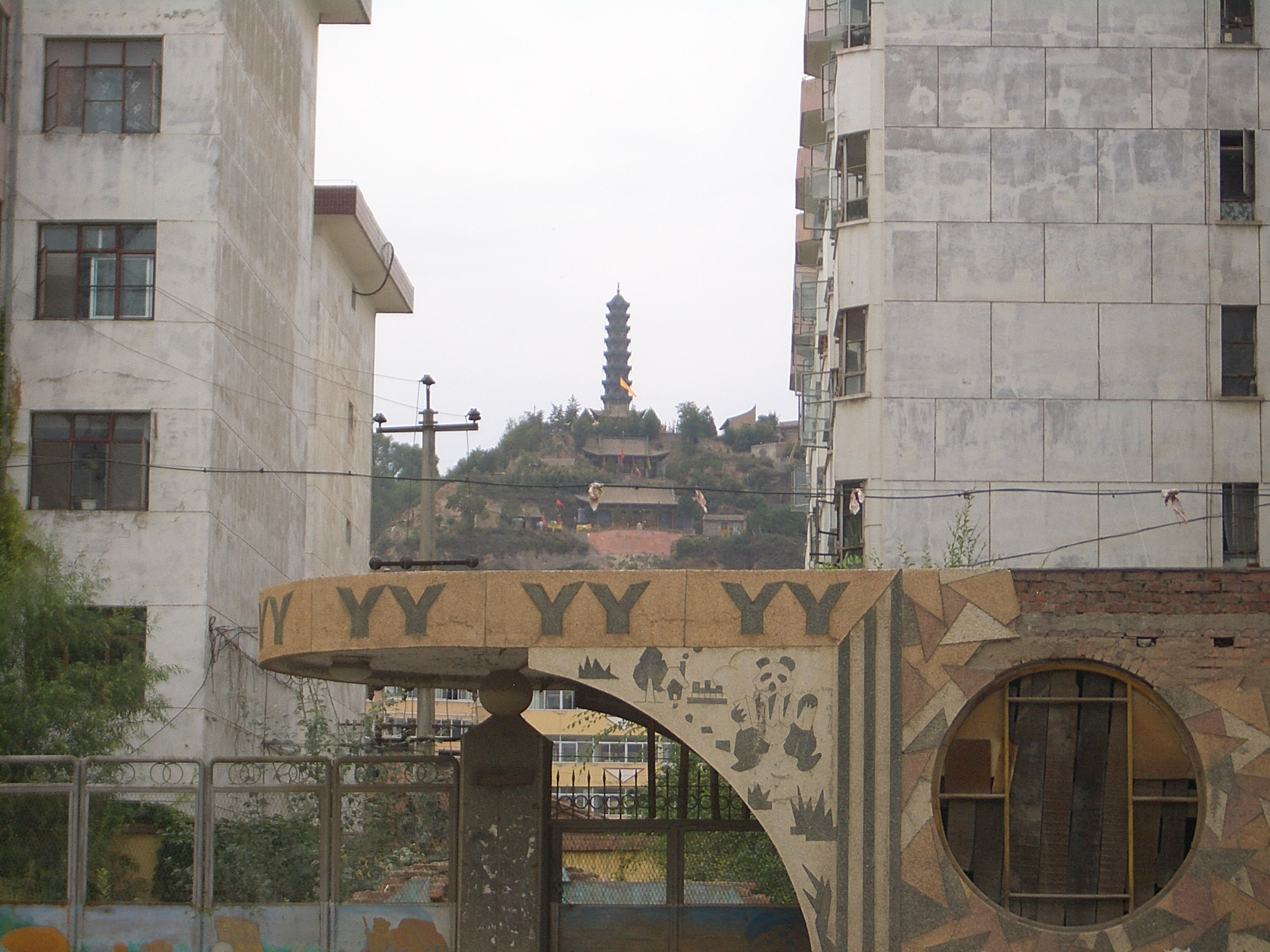}
  \caption{Complex skyline or structure}
\end{subfigure}\hfill
\begin{subfigure}[t]{0.3\textwidth}
  \centering
  \includegraphics[height=0.13\textheight]{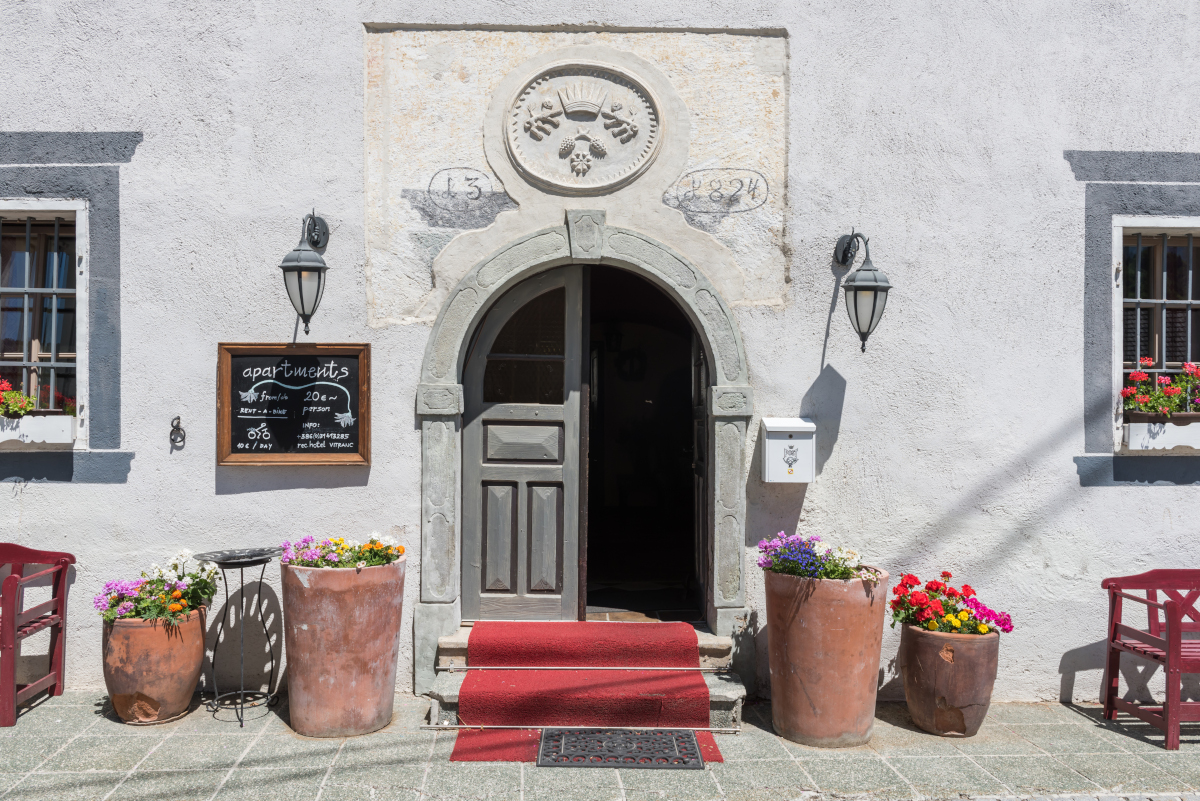}
  \caption{Non-monument with monument-like frontage}
\end{subfigure}\hfill
\begin{subfigure}[t]{0.3\textwidth}
  \centering
  \includegraphics[height=0.13\textheight]{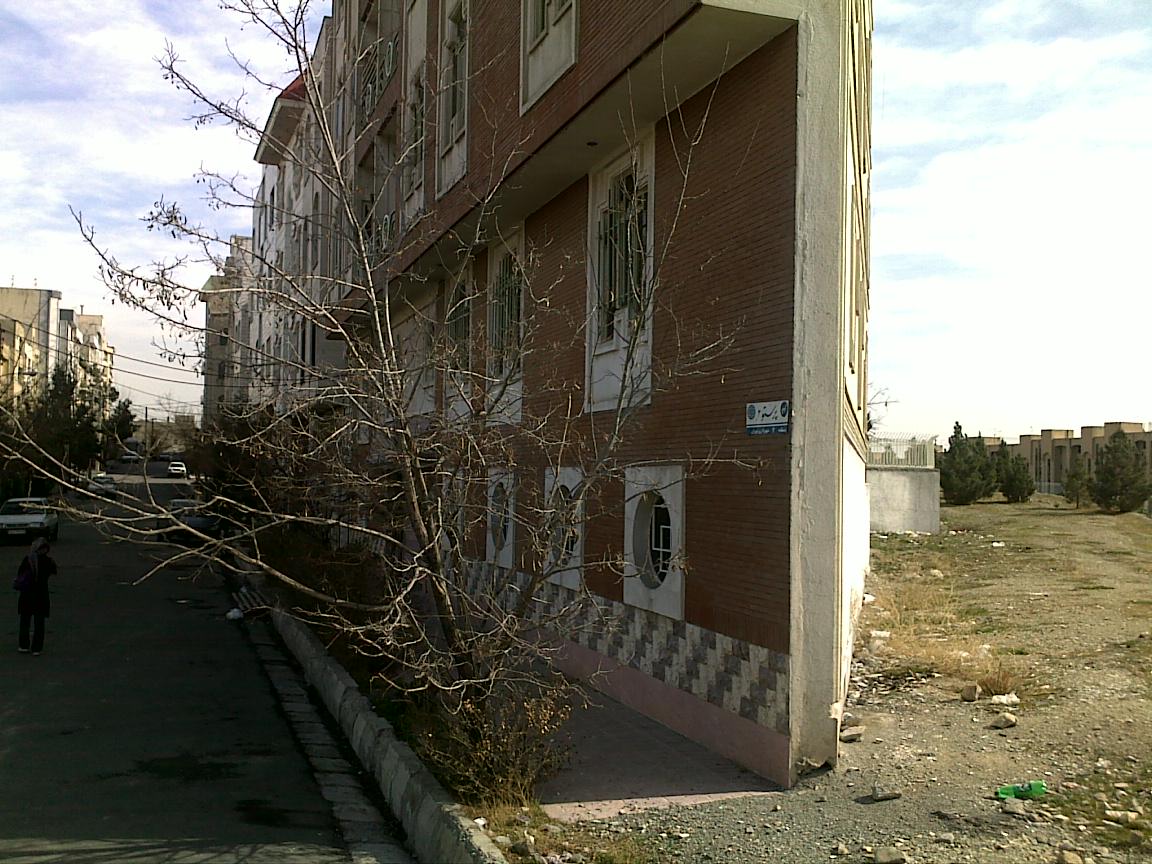}
  \caption{Architectural novelty or triangular geometry}
\end{subfigure}

\par\vspace{0.7em}

\textbf{Layer 3: known vs. unknown architectural style confusions at the near-OOD frontier.}\\[0.25em]

\begin{subfigure}[t]{0.3\textwidth}
  \centering
  \includegraphics[height=0.13\textheight]{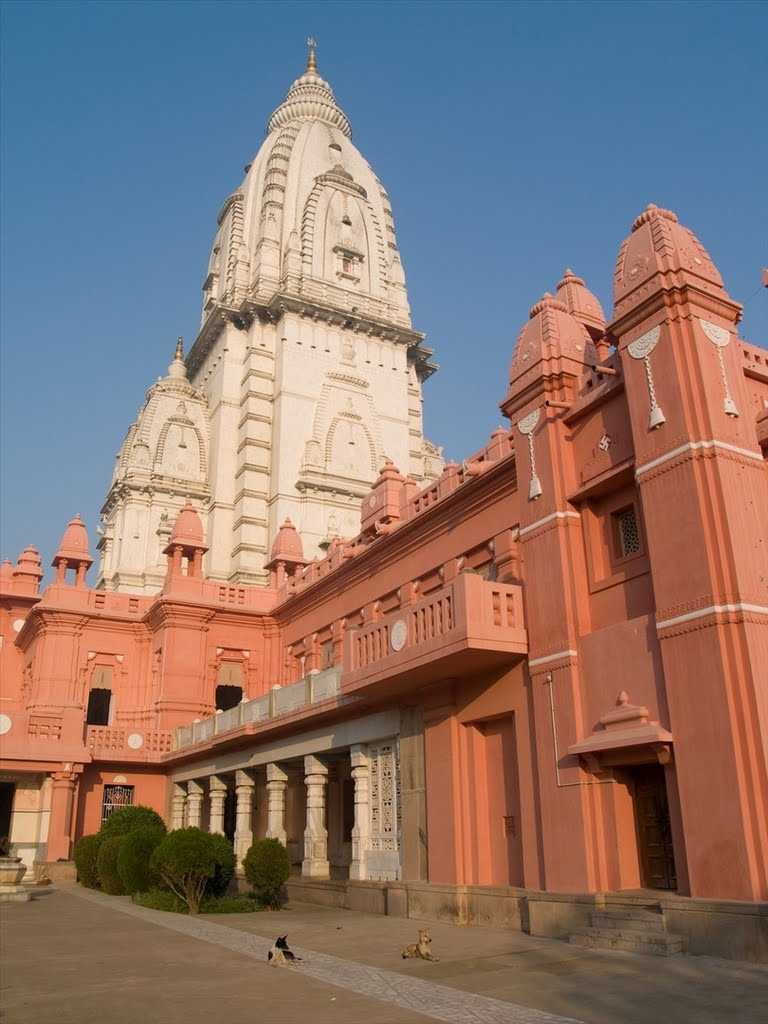}
  \caption{Unknown style bordering known classes}
\end{subfigure}\hfill
\begin{subfigure}[t]{0.3\textwidth}
  \centering
  \includegraphics[height=0.13\textheight]{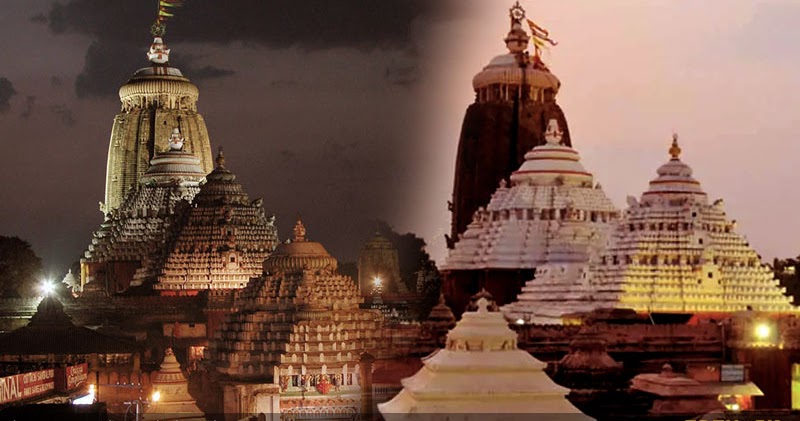}
  \caption{Mixed stylistic signals}
\end{subfigure}\hfill
\begin{subfigure}[t]{0.3\textwidth}
  \centering
  \includegraphics[height=0.13\textheight]{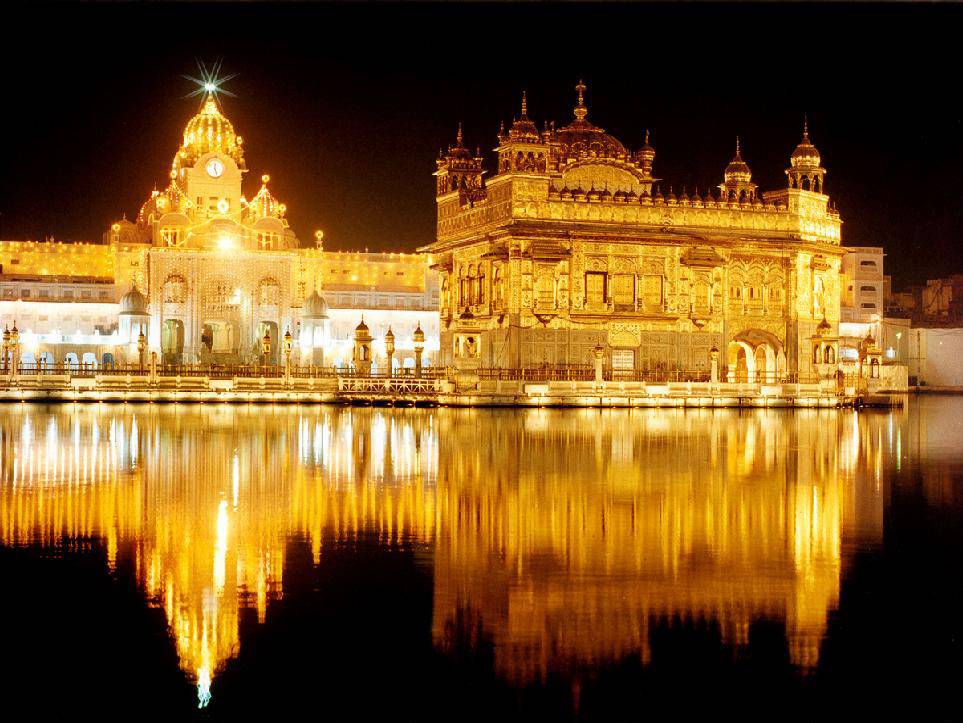}
  \caption{Degraded or atypical viewpoint}
\end{subfigure}

\caption{Qualitative examples of inputs misclassified at each filtering layer in SeNeDiF-OOD applied to MonuMAI.}
\label{fig:layerN_mis}
\end{figure}

For each layer, we include a three-image panel summarizing typical misclassifications. Figure~\ref{fig:layerN_mis} shows representative examples of the few inputs that confused the corresponding models. These visuals provide a compact qualitative view of the residual errors at each boundary and will be revisited later to extract common patterns and motivate targeted improvements.

\FloatBarrier

\subsubsection{Progressive narrowing along the cascade: visual walkthrough}

\begin{figure}[ht]
    \centering
    \includegraphics[width=\textwidth]{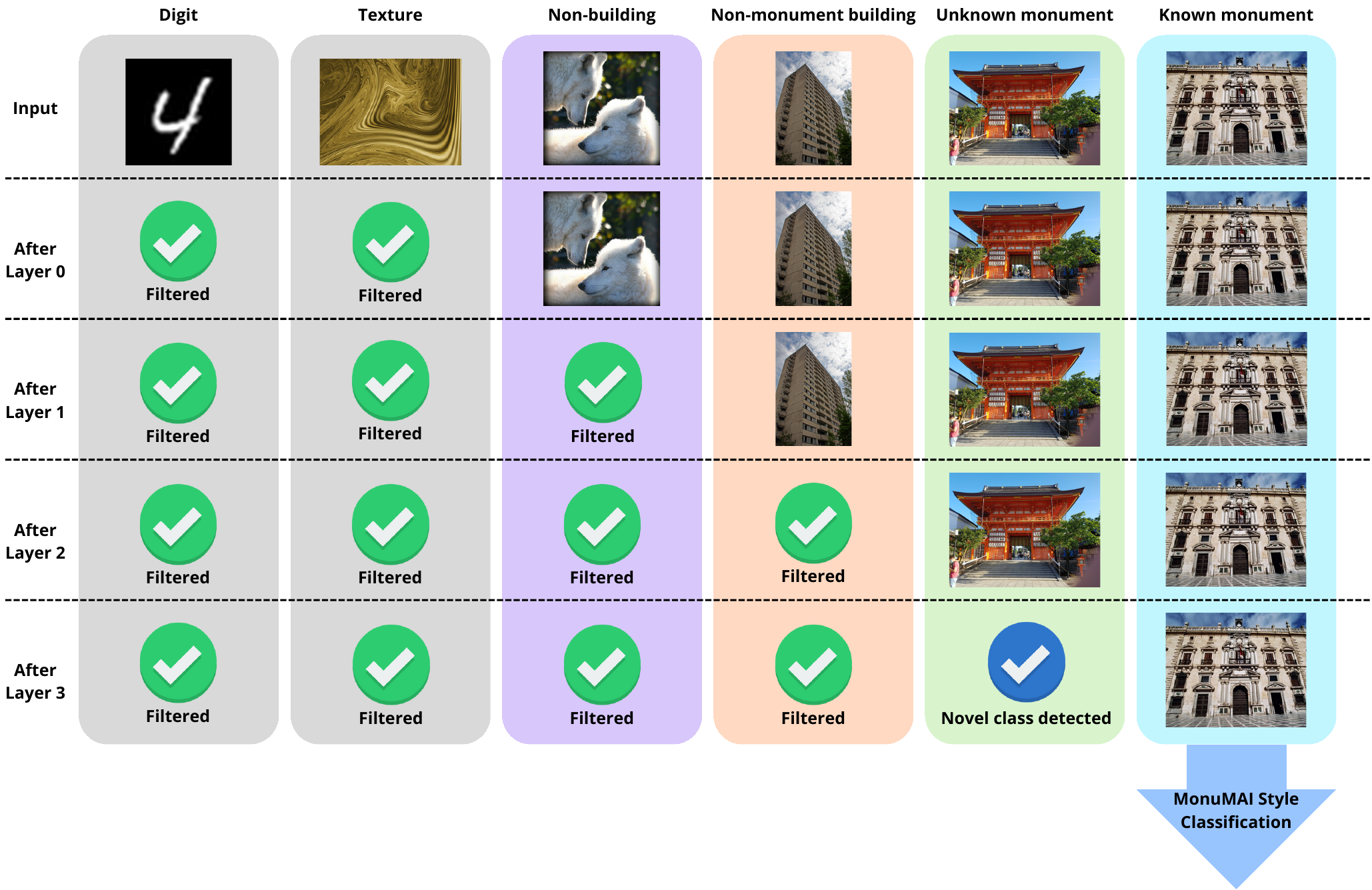}
    \caption{Cohort tracking across the SeNeDiF-OOD methodology applied to MonuMAI.}
    \label{fig:cohort_tracking}
\end{figure}

Figure~\ref{fig:cohort_tracking} illustrates the progressive narrowing of the input domain along the hierarchical fusion cascade. Each column fixes a real example from the MonuTest-OOD set (digit, texture, non-building, non-monument building, unknown monument, known monument), and each row shows the state after a given layer (\emph{Input}, \emph{After Layer 0–3}). The images are the actual test samples we exposed to our models during evaluation, not synthetic mock-ups. Empty cells mark the layer at which a sample is filtered out, while surviving thumbnails proceed to the next row. This compact view makes the semantics of each boundary explicit and complements the quantitative results reported earlier.

\FloatBarrier

\subsection{Discussion}\label{sec:discussion}

In this section, we analyze the results obtained from the different experiments conducted.

\subsubsection{Layer-wise insights}
\label{sec:disc-layers}

\begin{itemize}

\item \textbf{Layer~0 — far-world filtering.}
The outer gate enforces a coarse but vital constraint: inputs must be compatible with natural-image statistics and generic, real-world content. In practice, this removes clearly alien inputs at negligible cost for downstream recall. The misclassification panel in Fig.~\ref{fig:layerN_mis}, Layer 0 shows that residual errors are not random: families such as texture-like patterns can share low-level statistics with photographs and therefore require a more semantic check later in the cascade. This behaviour is by design. Rather than force a brittle decision at the periphery, the gate tolerates ambiguous near-world cases and defers them to task-aware filters where the relevant notions of “building” and “monument” can be applied.

\item \textbf{Layer~1 — building vs.\ non-building.}
This gate asserts scene plausibility with respect to the built environment. Its operating point is intentionally conservative: it favours specificity so that unmistakably non-architectural content is stopped early, protecting the inner stages (Table~\ref{tbl:layer1}). The residual errors (Fig.~\ref{fig:layerN_mis}, Layer 1) are instructive. Fountains, statue ensembles or even drawings that suggest façade-like structure can momentarily mimic building cues if perspective, materials or composition align. Conversely, a small fraction of genuine buildings with extreme viewpoints, heavy occlusion or atypical framing may be rejected. From a safety perspective, this bias is acceptable: occasional false rejects at this stage mainly affect edge cases with weak architectural evidence, whereas false accepts would propagate uncertainty deeper into the pipeline.

\item \textbf{Layer~2 — monument vs.\ building.}
Once the stream contains buildings, the question becomes architectural relevance. This is where the curated dataset assembled via Wikimedia Commons and CLIP-based thematic filtering is pivotal. The harvesting and multi-prompt screening provide breadth without drifting off-topic, which translates into balanced behaviour across subcategories (Tables~\ref{tbl:layer2_accuracy} and \ref{tbl:layer2}). The errors in Fig.~\ref{fig:layerN_mis}, Layer 2 reflect the hardest cases for this boundary: modern or anonymous buildings that exhibit monument-like cues (portals, axial symmetry, salient ornament), complex skylines where salient structures draw attention away from context, or geometric novelties that read as “significant” despite lacking historical or stylistic value. In practice, this gate functions as the workhorse for the near-world: it shields subsequent stages from generic buildings while retaining both known and unknown monuments for final adjudication.

\item \textbf{Layer~3 — known vs.\ unknown style.}
The innermost decision is a near-OOD problem within the monument domain: distinguish styles recognized by the application from the rest. Here, outlier exposure provides the right training signal: the classifier is encouraged to remain confident on the supported styles while assigning high uncertainty to stylistically out-of-scope monuments. The result is a clean semantic frontier that can be calibrated for deployment and explained to users in terms of the taxonomy they already see in the app (Table~\ref{tbl:layer3}; examples in Fig.~\ref{fig:layerN_mis}, Layer 3). Importantly, this gate does more than reject: it produces a structured set of “unknown-style” monuments that is well suited for human-in-the-loop review and future expansion of the taxonomy, closing the loop between OOD detection and active learning.

\end{itemize}

\medskip
Taken together, these layer-wise observations explain and complement the global behavior shown in Table~\ref{tbl:global_results}: the cascade distributes uncertainty to the stage where it can be handled with the most appropriate inductive bias, and the few residual mistakes are concentrated in well-understood visual regimes that suggest concrete avenues for refinement in subsequent sections. This progressive narrowing of the operational design domain is visually summarized in Fig. \ref{fig:cohort_tracking}, which tracks heterogeneous samples through the sequence of gates. The diagram externalizes the decision logic, showing precisely which semantic boundary (e.g., "non-building," "non-monument," "unknown style") rejects each OOD category. This traceability not only underpins the system's interpretability and robustness by deferring ambiguous decisions, but also makes each rejection actionable for targeted maintenance and principled taxonomy expansion.

\subsubsection{Global insights}

The end-to-end results, summarized in Table \ref{tbl:global_results}, demonstrate the cascade's effectiveness in rebalancing the system's error profile for open-world operation. The SeNeDiF-OOD pipeline attains high accuracy (0.9417) and strong separability (AUROC 0.9642). Notably, precision is high (0.8823), confirming that inputs accepted as known monuments are rarely false positives.

This high precision is achieved via a conservative design, reflected in a moderate recall (0.7625). The sequence of upstream gates and the final known-versus-unknown style boundary are calibrated to reject ambiguous cases rather than admit non-monuments or unfamiliar styles. This trade-off is the primary practical gain, as it directly mitigates the critical failure of the original MonuMAI pipeline. The legacy system, by prioritizing sensitivity (0.8991 recall), suffered from extremely low precision (0.4130), accepting a large volume of OOD content as false positives. The nested cascade resolves this, drastically improving precision while maintaining strong overall performance.

Beyond these aggregate gains, our methodology yields concrete operational benefits. First, it reduces noise by filtering inputs at the earliest compatible boundary, lowering server load and stabilizing the end-to-end experience. Second, the structure converts rejection into traceable information; each gate corresponds to a user-comprehensible concept (e.g., "non-building" or "unknown style"), making failures actionable. Third, maintenance becomes modular, as thresholds and data sources can be tuned per-layer without destabilizing the system. Finally, the SeNeDiF-OOD pipeline creates a principled pathway for growth, as images withheld by the style gate (Layer 3) form a high-quality pool for expert curation and active learning. In short, the cascade organizes the open world for reliable operation today and coherent evolution tomorrow.

\subsubsection{Error typology and visual patterns} \label{sec:disc-errors}

A small number of recurring visual regimes account for most residual mistakes across the cascade. In \autoref{fig:layerN_mis}, we group them into patterns aligned with the semantic boundaries enforced by each layer, which clarifies both \emph{why} they arise and \emph{where} they should be addressed.

\begin{itemize}

\item \textbf{Texture-like statistics:} Images dominated by fine, repetitive patterns may share low-level statistics with natural photographs, allowing them to pass Layer 0. They are correctly deferred to Layer 1, which can identify the lack of scene-level semantics (e.g., vanishing points or volumetric layout) proper to a built environment.

\item \textbf{Pseudo-architectural content:} Urban features like fountains, statue ensembles, or scaffolds can momentarily mimic façade structures and slip past Layer 1 (\autoref{fig:layerN_mis}, Layer 1). These are precisely the inputs Layer 1 is meant to arbitrate, and the few that pass are scrutinized by Layer 2, which requires true architectural \emph{relevance}, not just built appearance.

\item \textbf{Generic buildings with monument-like cues:} This is the hardest near-OOD boundary, where modern or anonymous buildings present features associated with monumental imagery, such as axial symmetry or salient portals (\autoref{fig:layerN_mis}, Layer 2). The breadth of the CLIP-curated corpus is pivotal here, mitigating this by exposing the classifier to diverse non-monument negatives.

\item \textbf{Degraded or atypical Inputs:} Heavy blur, low light, aggressive crops, or novel geometries (e.g., triangular houses) reduce the availability of diagnostic cues. The cascade behaves conservatively, rejecting these inputs at the earliest boundary where evidence is insufficient (typically Layer 1 or 2), which is consistent with the safety objective. 

\item \textbf{Mixed or borderline stylistic signals:} Within the monument domain, façades combining elements from multiple traditions (\autoref{fig:layerN_mis}, Layer 3) represent the near-OOD frontier for Layer 3. Outlier exposure correctly encourages high uncertainty on these inputs, producing a semantically meaningful "unknown-style" pool for human curation.

\item \textbf{Context suppression:} A transversal factor where tight crops or foreground clutter hide situational cues (e.g., surrounding urban fabric, scale). This can confuse Layer 1 (fountain vs. façade) or Layer 2 (landmark vs. office), making decisions more brittle.
\end{itemize}

\medskip Overall, these patterns confirm that residual mistakes are not arbitrary outliers but are concentrated in well-defined visual regimes aligned with the cascade’s semantic boundaries. This alignment is precisely what makes the SeNeDiF-OOD pipeline maintainable: each regime suggests a targeted remedy at the layer where the relevant concept is enforced, turning the error typology into a roadmap for iterative refinement.


\section{Conclusions} \label{sec:conclusions}


In this work, we have introduced SeNeDiF-OOD, a novel methodology that operationalizes OOD detection and open-world recognition by integrating semantic nested dichotomies into a robust decision fusion architecture. The approach is motivated by the inherent open-world setting of applications such as MonuMAI, where user-submitted inputs frequently extend beyond the intended scope of the model. Our contribution has been twofold. At the theoretical level, we have formalized a taxonomy of OOD categories and established a general framework for applying nested dichotomies to image-based OOD detection. At the practical level, we have implemented this methodology within the MonuMAI pipeline, designing and training dedicated filtering layers that progressively refine the decision boundaries.

The experimental analysis demonstrates that SeNeDiF-OOD significantly improves the robustness of the MonuMAI pipeline, providing finer-grained filtering of irrelevant inputs while maintaining the correct preservation of in-distribution images. In particular, the layer-by-layer evaluation highlights the complementary role of each stage and the resulting fusion of their decisions, from coarse rejection of far-world images to the identification of architectural styles not present in the training data. The completed experiments corroborate this pattern, showing how the hierarchical decomposition distributes the decision burden and yields a clearer separation of responsibilities across layers.

Furthermore, the hierarchical design offers additional benefits beyond raw performance, as the layered filtering process enables an explicit fusion of semantic evidence across dichotomy layers. First, the layered filtering process provides enhanced interpretability, as each rejection can be traced back to the corresponding dichotomy layer, offering insights into the nature of the input and enabling targeted diagnosis. Second, the ability of the final layer to detect novel classes paves the way for integrating active learning strategies, where new styles can be gradually incorporated into the system. Finally, the approach contributes to responsible AI by promoting robustness, explainability, and alignment with human reasoning in open-world environments, while also improving auditability and maintainability of the overall pipeline.

Despite these advances, several limitations remain. The current implementation depends on carefully curated datasets at each stage, which may limit scalability when transferring the framework to other domains or when the data distribution evolves. The methodology also relies on periodic calibration and threshold selection, which should be monitored to avoid drift under changing conditions. In addition, certain challenging cases, such as images with severe degradation, atypical viewpoints, or borderline semantics can still lead to conservative rejections that, while desirable from a safety perspective, merit further attention through data augmentation and curation.

Future work will address these limitations by exploring adaptive training strategies for each dichotomy layer, incorporating more diverse sources of outlier exposure, and extending the framework to additional cultural heritage datasets. We also aim to develop active learning loops with human-in-the-loop supervision so that the system can identify and progressively integrate new architectural styles. Finally, improving efficiency through model compression, cascade design, and lightweight on-device inference, together with continuous calibration and monitoring, will be key to ensuring practical and sustainable deployment.

This work lays the foundation for a new line of research in OOD detection for real-world computer vision systems through nested dichotomies, providing both theoretical grounding and a practical fusion of methodology and deployment on a challenging and socially relevant task such as monumental heritage classification. The decomposition into semantically meaningful decisions not only strengthens robustness but also turns rejections into actionable information, enabling a safer and more interpretable pathway for open-world recognition. 

In summary, our work validates the central thesis: decomposing OOD detection into semantically meaningful binary decisions yields a safer, clearer, and more maintainable system. The SeNeDiF-OOD not only improves quantitative performance, especially in the near-OOD frontier, where errors are the most subtle, but also turns rejections into information through decision fusion: each \emph{no} says \emph{why}. Beyond the specific case of MonuMAI, this design principle can be instantiated in other real-world computer vision pipelines that must operate under open-world uncertainty and strict safety requirements. We expect these results to foster further research at the intersection of OOD detection, hierarchical decision making, and responsible deployment of vision models in multiple domains. 

\subsection*{Acknowledgements}

This research was supported by the Strategic Project IAFER-Cib [C074/23] through a collaboration agreement signed between the National Institute of Cybersecurity (INCIBE) and the University of Granada. This initiative is carried out within the framework of the Recovery, Transformation and Resilience Plan funds, financed by the European Union (Next Generation EU).

\bibliographystyle{ieeetr}
\bibliography{references.bib}

\end{document}